\documentclass[runningheads]{llncs}

\usepackage[mobile]{eccv}

\usepackage{eccvabbrv}

\usepackage{graphicx}
\usepackage{booktabs}

\usepackage[accsupp]{axessibility}  % Improves PDF readability for those with disabilities.

\usepackage{booktabs, multirow} % for borders and merged ranges
\usepackage{soul}% for underlines
\usepackage{adjustbox}
\usepackage{changepage,threeparttable} % for wide tables

\usepackage{url}

\usepackage[utf8]{inputenc} % allow utf-8 input
\usepackage[T1]{fontenc}    % use 8-bit T1 fonts
\usepackage{amsfonts}       % blackboard math symbols
\usepackage{nicefrac}       % compact symbols for 1/2, etc.
\usepackage{microtype}      % microtypography
\usepackage{xcolor}         % colors
\usepackage[linesnumbered,ruled,vlined]{algorithm2e}
\usepackage{array}
\usepackage{amssymb}
\usepackage{latexsym}
\usepackage{amsmath,amssymb}
\usepackage{graphicx}
\usepackage{multirow}
\usepackage{adjustbox}
\usepackage{makecell}
\usepackage{graphicx}
\usepackage{kotex}
\usepackage{booktabs, multirow} % for borders and merged ranges
\usepackage{nicematrix,caption}
\usepackage{soul}% for underlines
\usepackage{changepage,threeparttable} % for wide tables
\usepackage{subcaption}
\usepackage{wrapfig}
\usepackage[
linesnumbered,ruled,vlined]{algorithm2e}
\usepackage {algpseudocode}
\usepackage{algorithmicx}
\usepackage{algcompatible}
\usepackage{bm}
\usepackage[pagebackref,breaklinks,colorlinks,citecolor=eccvblue]{hyperref}

\usepackage{orcidlink}
\newcommand*{\V}[1]{\mathbf{#1}}
\newcommand*{\C}[1]{\mathcal{#1}}
\newcommand{\bluetext}[1] {\textcolor{blue}{#1}} % for revision

\begin{document}

\title{BeyondScene: Higher-Resolution Human-Centric Scene Generation With Pretrained Diffusion}

\titlerunning{BeyondScene}

\author{Gwanghyun Kim\inst{1\star},
Hayeon Kim\inst{1\star},
Hoigi Seo\inst{1\star},
Dong Un Kang\inst{1}\thanks{Authors contributed equally. \ $^{\dagger}$ Corresponding author.}, \\
Se Young Chun\inst{1,2}$^{\dagger}$}

\authorrunning{Kim et al.}

\institute{$^1$Dept. of Electrical and Computer Engineering, $^2$INMC \&  IPAI  \\
Seoul National University, Republic of Korea \\
\email{\{gwang.kim, qkrtnskfk23, seohoiki3215, khy5630,  sychun\}@snu.ac.kr}   } 

\maketitle

\begin{abstract}
  Generating higher-resolution human-centric scenes with details and controls remains a challenge for existing text-to-image diffusion models. This challenge stems from limited training image size, text encoder capacity (limited tokens), and the inherent difficulty of generating complex scenes involving multiple humans. While current methods attempted to address training size limit only, they often yielded human-centric scenes with severe artifacts. We propose BeyondScene, a novel framework that overcomes prior limitations, generating exquisite higher-resolution (over 8K) human-centric scenes with exceptional text-image correspondence and naturalness using existing pretrained diffusion models. BeyondScene employs a staged and hierarchical approach to initially generate a detailed base image focusing on crucial elements in instance creation for multiple humans and detailed descriptions beyond token limit of diffusion model, and then to seamlessly convert the base image to a higher-resolution output, exceeding training image size and incorporating details aware of text and instances via our novel instance-aware hierarchical enlargement process that consists of our proposed high-frequency injected forward diffusion and adaptive joint diffusion. BeyondScene surpasses existing methods in terms of correspondence with detailed text descriptions and naturalness, paving the way for advanced applications in higher-resolution human-centric scene creation beyond the capacity of pretrained diffusion models without costly retraining. Project page: \url{https://janeyeon.github.io/beyond-scene}.
\end{abstract}

% {
%   \renewcommand{\thefootnote}%
%     {\fnsymbol{footnote}}
%     \footnotetext[1]{Authors contributed equally. \ $^\dagger$Corresponding author.}
%     % \footnotetext[1]{Authors contributed equally.}
%     % \footnotetext[2]{Corresponding author.}
  
% }

\begin{figure}[!t]
    \centering
    \includegraphics[width=\textwidth]{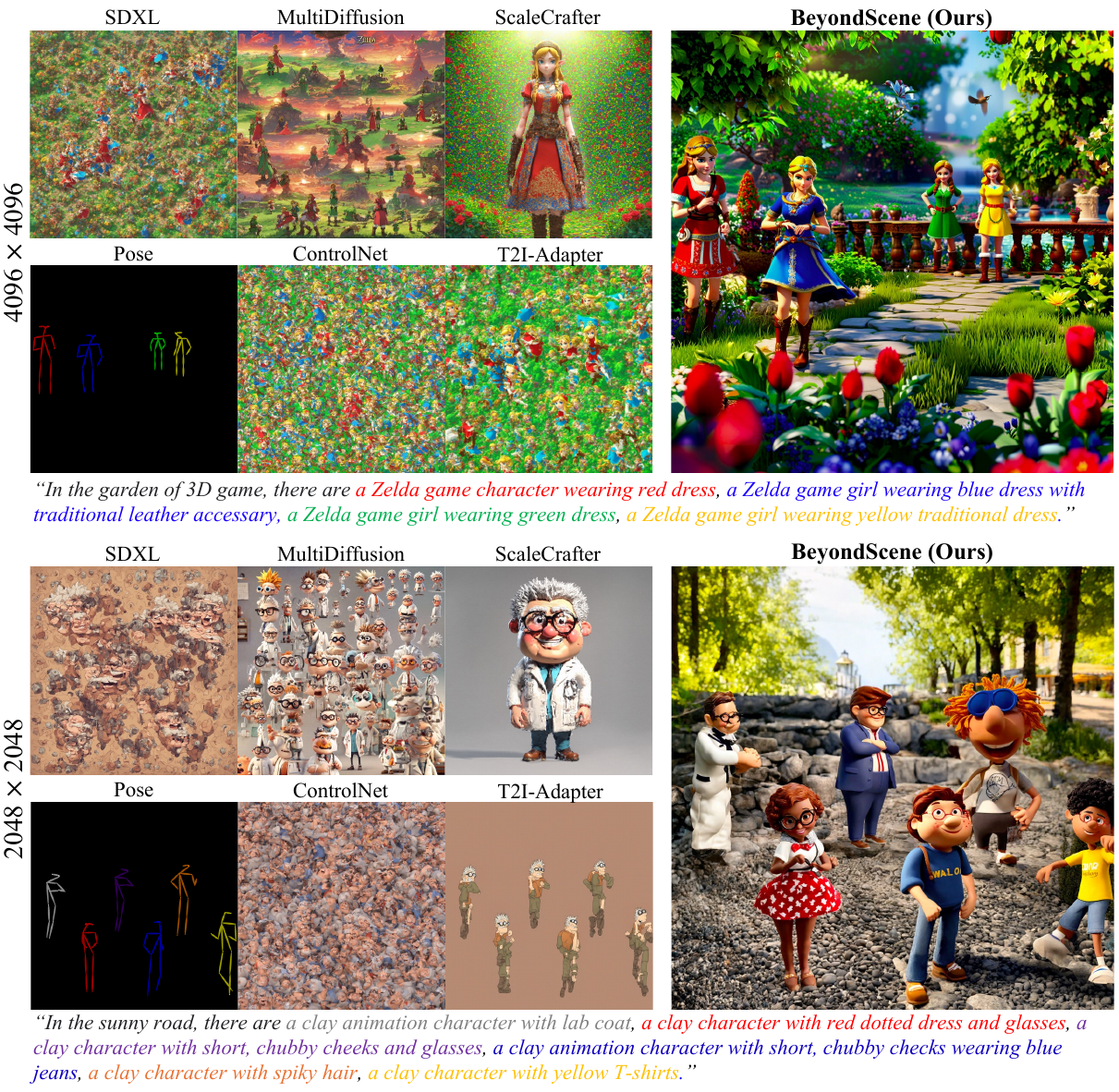}
   \vspace{-2em}
    \caption{BeyondScene pushes the boundaries of high-resolution human-centric scene generation. Unlike existing methods that often suffer from unrealistic scenes, anatomical distortions, and limited text-to-image correspondence, BeyondScene excels in 1) highly detailed scenes, 2) natural and diverse humans, 3) fine-grained control. This breakthrough paves the way for groundbreaking applications in human-centric scene design.  The color in each description represents the description for each instance that has the same color in the pose map.}
    \label{fig_quali_custom1}
   \vspace{-2em}
\end{figure}

\section{Introduction}
\label{sec:intro}
Human-centric scene generation~\cite{poseguided17_gan,decomposed20_gan,mustgan21_gan,pise21_gan,spgnet21_gan,ctnet21_gan,nted22_gan,dptn22_gan, transformation20_vae,finegrained21_vae,kpe22_vae,
verbalpersonnets22,sharedspace21,tips22, controlnet23, gligen23, ju2023humansd, t2i23, saharia2022photorealistic}, encompassing the creation of images featuring individuals under specified conditions, has emerged as a critical research area with significant academic and industrial applications. Its potential extends beyond animation and game production to comprise diverse domains. 

While recent advancements utilizing text-to-image (T2I) diffusion models~\cite{ramesh2022hierarchical,midjourney,rombach2022highresolution} have yielded promising results in generating controllable human scenes~\cite{controlnet23, gligen23, ju2023humansd, t2i23}, scaling these methods to handle larger and more complex scenes such as involving multiple humans remains a significant challenge. These limitations primarily stem from following causes. 1) \textit{Training image size}: Directly sampling from a limited training image size can introduce artifacts and constrain the final scene resolution. 2) \textit{Limited text encoder tokens}: The restricted number of tokens in text encoders (typically 77 for Stable Diffusion) hinders the inclusion of detailed descriptions for multiple instances within the scene. 3) \textit{Inherent limitations of T2I diffusion models}: These models struggle to generate complex scenes with several human figures and intricate details.

Existing attempts to address the training size issue, such as joint-process~\cite{bar2023multidiffusion, zhang2023diffcollage, lee2023syncdiffusion} and dilation-based methods~\cite{he2023scalecrafter}, only partially address the problem. These methods often introduce new challenges specific to human-centric scenes as represented in Fig.~\ref{fig_quali_custom1}, including:
1) \textit{Unrealistic scenes and objects}: These methods can generate nonsensical scenarios with duplicated objects, humans defying gravity, and physically distorted environments.
2) \textit{Anatomical distortions}: Generated scenes may exhibit unrealistic human anatomy, such as abnormal limbs or facial features.
3) \textit{Limited correspondence}: Existing methods often fail to capture the complexity of scenes with multiple human instances and detailed descriptions, often generating single objects and lacking control over specific details like clothing or hairstyles.

We propose BeyondScene, a novel framework that overcomes these limitations, generating high-resolution (over 8K) human-centric scenes with exceptional text-image correspondence and naturalness. BeyondScene employs a staged and hierarchical approach, which looks similar to a classical multi-resolution manner, but it is closer to how artists establish a foundation before adding details.

Firstly, BeyondScene generates a \textit{detailed base image} focusing on essential elements, human poses and detailed descriptions beyond the token limits. This initial stage enables detailed instance creation for multiple humans, surpassing the limitations of text encoders. Secondly, the method leverages an \textit{instance-aware hierarchical enlargement} process to convert this base image to a higher resolution output beyond the training image size. Unlike naive super-resolution methods that simply scale resolution without considering text and instances, our approach refines content and adds more details aware of text and instances. This is achieved through our proposed novel techniques, including 1) \textit{high frequency-injected forward diffusion} that addresses the issue of blurred low-quality results in image-to-image translation with the upsampled image by adaptively injecting high-frequency details into the upsampled image, enhancing the final output while preserving content and 2) \textit{adaptive joint diffusion} that facilitates efficient and robust joint diffusion while maintaining control over human characteristics like pose and mask. This approach utilizes view-wise conditioning of text and pose information, along with a variable stride for the joint process. Furthermore, similar to how artists add details progressively, our method allows for the addition of details at different stages, exploiting the changing receptive field of each view.

\begin{figure}[!t]
    \centering
    \includegraphics[width=\textwidth]{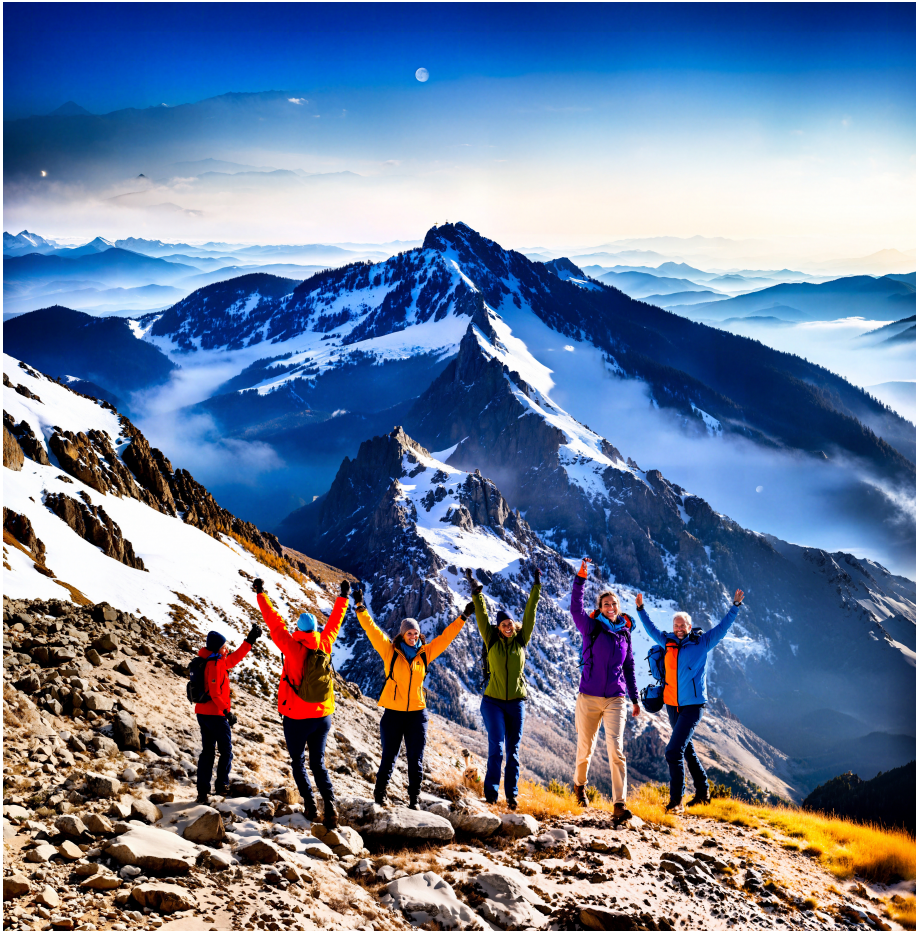}
   \vspace{-1.5em}
    \caption{\textbf{Beyond 8K ultra-high resolution image}. This 8192$\times$8192 image, generated by BeyondScene, surpasses the training resolution of SDXL by 64$\times$, while exceeding the technical classification of 8K (7680$\times$4320).}        
    \label{fig_quali_8k}
   \vspace{-2em}
\end{figure}

We comprehensively evaluate BeyondScene using qualitative and quantitative metrics, along with user studies. The results demonstrate significant improvements over existing methods in terms of 1) correspondence with detailed text descriptions and 2) naturalness and reduction of artifacts. 
Furthermore, we showcase the result of 8192$\times$8192 image generation beyond 8K ultra high-resolution, demonstrating the capability for generating even higher-resolution images as displayed in Fig.~\ref{fig_quali_8k}.  
These advancements pave the way for exciting new applications in higher-resolution human-centric scene creation.

\section{Related work}
\subsection{Controllable Human Generation}

Early approaches to controllable human generation~\cite{poseguided17_gan,decomposed20_gan,mustgan21_gan,pise21_gan,spgnet21_gan,ctnet21_gan,nted22_gan,dptn22_gan, transformation20_vae,finegrained21_vae,kpe22_vae} relied on pose guidance and source images, achieving success within specific scenarios but struggling with diverse scenes and arbitrary poses. Text-based conditioning methods emerged with the rise of large vision-language models~\cite{clip}, but initial attempts~\cite{verbalpersonnets22,sharedspace21,tips22} faced limitations in vocabulary size and open vocabulary settings. Recent T2I diffusion~\cite{ramesh2022hierarchical,midjourney,rombach2022high,podell2023sdxl}-based methods like ControlNet~\cite{controlnet23}, T2I-Adapter~\cite{t2i23}, GLIGEN~\cite{gligen23}, and HumanSD~\cite{ju2023humansd} introduced methods for incorporating diverse conditions, but scaling to larger scenes with multiple individuals remains a challenge. This work proposes BeyondScene, a novel framework specifically designed to overcome these limitations and enable the generation of high-resolution human-centric scenes.

\subsection{Large Scene Generation Using Diffusion Models}

Achieving high-resolution image generation~\cite{ding2023patched, ramesh2022hierarchical} presents significant hurdles. Training models from scratch or fine-tuning pretrained models~\cite{ramesh2022hierarchical,midjourney,rombach2022high,podell2023sdxl, zhang2022styleswin, rombach2021highresolution, teng2023relay, hoogeboom2023simple, chen2023importance, zheng2023any, xie2023difffit} requires immense computational resources and struggles with the complexity of high-dimensional data. Recent exploration has ventured into training-free methods, with approaches like MultiDiffusion~\cite{bar2023multidiffusion}, SyncDiffusion~\cite{lee2023syncdiffusion}  (joint diffusion), and ScaleCrafter~\cite{he2023scalecrafter}  (dilation-based) emerging. However, these methods often introduce challenges specific to human-centric scenes, including unrealistic objects, anatomical distortions, and limited correspondence between text descriptions and generated images.
BeyondScene, our proposed framework, addresses these limitations through detailed base image generation and instance-aware hierachical enalargement.

\section{BeyondScene}
\begin{figure}[!t]
    \centering
    \includegraphics[width=0.99\textwidth]{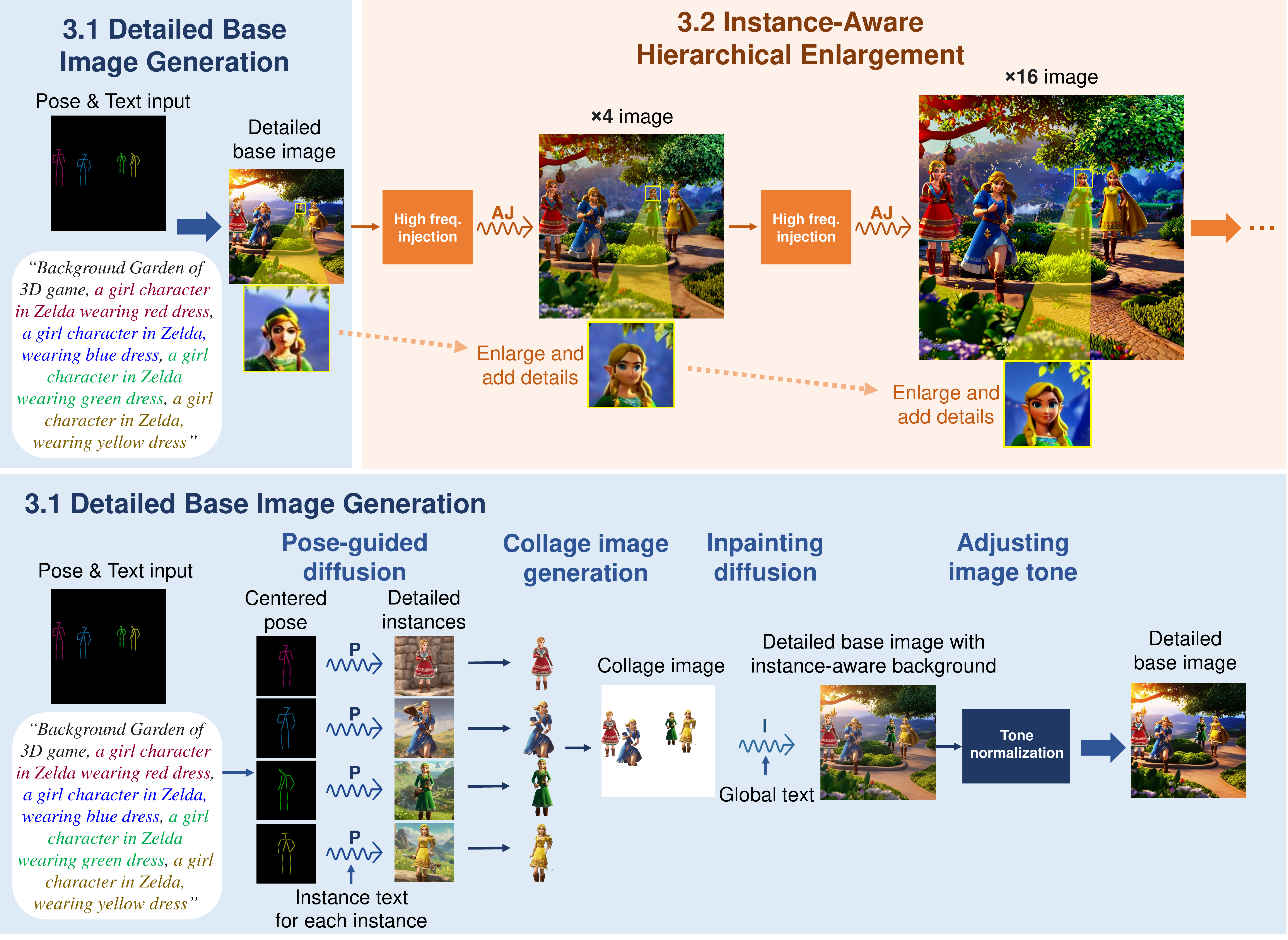}
   \vspace{-1em}
    \caption{BeyondScene generates high-resolution images in two stages. First, individual instances are created using pose-guided T2I diffusion models, segmented, cropped, and placed onto an inpainted background. The tone of image is then normalized. In the second stage (illustrated in Fig.~\ref{fig_method2}), this base image is progressively enlarged while maintaining detail and quality, effectively refining the image and adding further details leveraging high-frequency injected forward diffusion and adaptive joint diffusion (AJ).}  
    \label{fig_method1}
   \vspace{-1em}
\end{figure}

Current approaches to generating large human-centric scenes typically attempt to fill the entire canvas at once, often leading to challenges in capturing complex details and ensuring physical realism. Inspired by the artistic workflow of human painters, who establish a foundation with key elements and progressively refine them, we introduce BeyondScene, a novel framework that generates high-resolution human-centric images through a staged and hierarchical approach as represented in Fig.~\ref{fig_method1}. 

BeyondScene operates in two key stages, 1) \textit{detailed base image generation} (Sec.~\ref{base_image_generation}) that focuses on generating high-quality base images with precise control over crucial elements like human poses and detailed descriptions beyond typical token limitations and 2) \textit{instance-aware hierarchical enlargement} (Sec.~\ref{hierachical_enlargement}) where the generated base image is progressively enlarged while maintaining control and adding scene details. 

The staged approach enables precise control over the generated content beyond the limitations of training image size and token restrictions. Also, by iteratively refining details, BeyondScene produces high-quality human-centric scenes with exceptional realism. 

\subsection{Detailed Base Image Generation}
\label{base_image_generation}
Generating human-centric scenes with multiple individuals using T2I diffusion models faces a key challenge: the limited number of tokens available in the text encoder restricts detailed descriptions to a few key elements. To overcome this, we generate detailed base image as illustrated in Fig.~\ref{fig_method1}.
Firstly, individual instances are generated in the training resolution of pose-guided T2I diffusion models~\cite{controlnet23, t2i23, ju2023humansd, gligen23}. This allows for detailed descriptions beyond token limitations, tailored to each individual. Subsequently, the generated instances are segmented, cropped, and resized to the desired size.
Next, a diffusion-based inpainting method is employed to create an instance-aware background that seamlessly integrates with the individual instances. This ensures a realistic and coherent background tailored to the specific scene.
To address potential inconsistencies in brightness and color tones across individual elements, tone normalization through contrast limited adaptive histogram equalization (CLAHE) is performed. 
This involves first creating a grid histogram for pixel values within a specified grid, followed by adjusting the local contrast to align with the image's overall contrast. Notably, direct equalization in the presence of noise within the grid could induce the noise amplification problem. To prevent this, equalization is preceded by a redistribution process that includes limiting the maximum height of the grid histogram. This procedure aims to achieve a visually cohesive and appealing base image. 
Through this approach, BeyondScene lays the foundation for a finely detailed base image that effectively captures the intricacies of the desired human-centric scene.

\begin{figure}[!t]
    \centering
    \includegraphics[width=\textwidth]{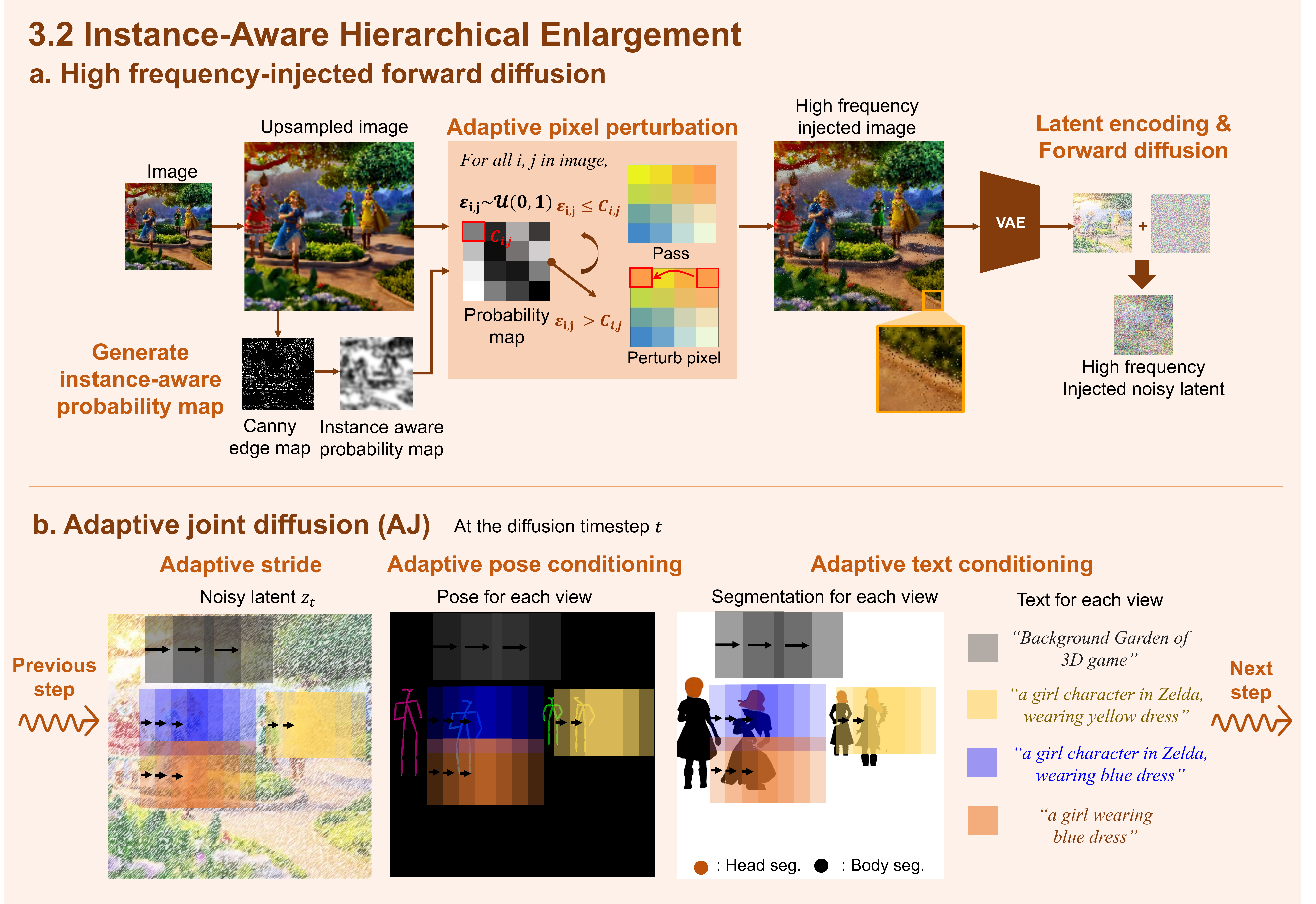}
   \vspace{-2em}
    \caption{Our instance-aware hierarchical enlargement involves two crucial processes: 1) High frequency-injected forward diffusion, which enables to achieve high resolution through a joint diffusion employing adaptive pixel perturbation. 2) Adaptive joint diffusion, dynamically regulating stride and conditioning of pose and text based on the presence of instances.}        
    \label{fig_method2}
   \vspace{-2em}
\end{figure}

\subsection{Instance-Aware Hierarchical Enlargement}
\label{hierachical_enlargement}

BeyondScene introduces a novel approach to address the challenge of generating high-resolution outputs beyond the training image size through a instance-aware hierarchical enlargement process, which surpasses mere scaling by actively refining content and incorporating additional details aware of text and instance.

In the initial step, a low-resolution image is upsampled, and a forward diffusion process is applied to create a noisy upsampled image. Subsequently, a joint diffusion process refines the image to a high resolution over the training resolution.
However, a drawback of this naive strategy is the induction of blurriness due to the model assigning sufficiently high likelihood to low-quality outputs. Also, this strategy often encounters issues such as unnatural duplication of objects and humans, stemming from the uniform application of the same text prompt to all views without pose conditioning.
To mitigate these, the paper proposes a novel \textit{high frequency-injected forward diffusion} and \textit{adaptive joint diffusion} as illustrated in Fig.~\ref{fig_method2}.

\subsubsection{High frequency-injected forward diffusion}
High frequency-injected forward diffusion enhances the translation of noisy latents from the upsampled image to high resolution with intricate details. This is achieved through a joint diffusion process employing adaptive pixel perturbation. 
This technique injects high-frequency details into the upsampled image based on a Canny edge map. This process enhances the final output's quality while preserving content.
While simply adding pixel noise sharpens the image, it can also introduce flickering artifacts at the borders. Our adaptive approach based on the Canny map significantly improves image quality and reduces these artifacts.
Specifically, we first up-sample the low-resolution image and calculate Canny map. Then Gaussian blur is applied to the map and normalized it to get probability map $C$. $\epsilon_{i, j} \sim \mathcal{U}(0, 1)$ is sampled on each pixel, $I_{i, j}$ which denotes $i$-th row and $j$-th column pixel of image $I$. If $\epsilon_{i, j} > C_{i, j}$, the pixel value within $d_{r}$ pixel distance is replaced so that we could apply pixel perturbation adaptively to avoid flickering artifacts near borders.

\subsubsection{Adaptive joint diffusion}
To mitigate duplication of objects and humans in the naive joint diffusion, we introduce adaptive conditioning and adaptive stride. 
\paragraph{Adaptive conditioning}
We introduce adaptive view-wise conditioning. This strategy leverages the segmentation maps obtained from the base image generation stage. For each view, we check which human instances are present. If an instance is included, we add its pose and detailed description to the global prompt used for that view in the diffusion model. Additionally, if the detailed text description includes specific parts (head, face, upper body, etc.), we can apply these details to the corresponding regions using fine-grained segmentation maps. This approach facilitates efficient and robust joint diffusion while maintaining control over crucial human characteristics like pose and appearance. Essentially, each view incorporates relevant text and pose information specific to the contents.

\paragraph{Adaptive stride}
To effectively enhance the quality of the generated scene, we employ an adaptive stride for the joint process. We reduce the stride in regions containing humans, ensuring that these areas capture fine details. In contrast, the stride is increased in background regions, allowing for more efficient computation in these areas.
This combination of adaptive view-wise conditioning and adaptive stride allows BeyondScene to effectively handle scenes with multiple humans while maintaining high-resolution detail and controllable image generation.
The details and algorithms on our methods are presented in the supplementary material.

\begin{figure}[!t]
    \centering
    \includegraphics[width=0.99\textwidth]{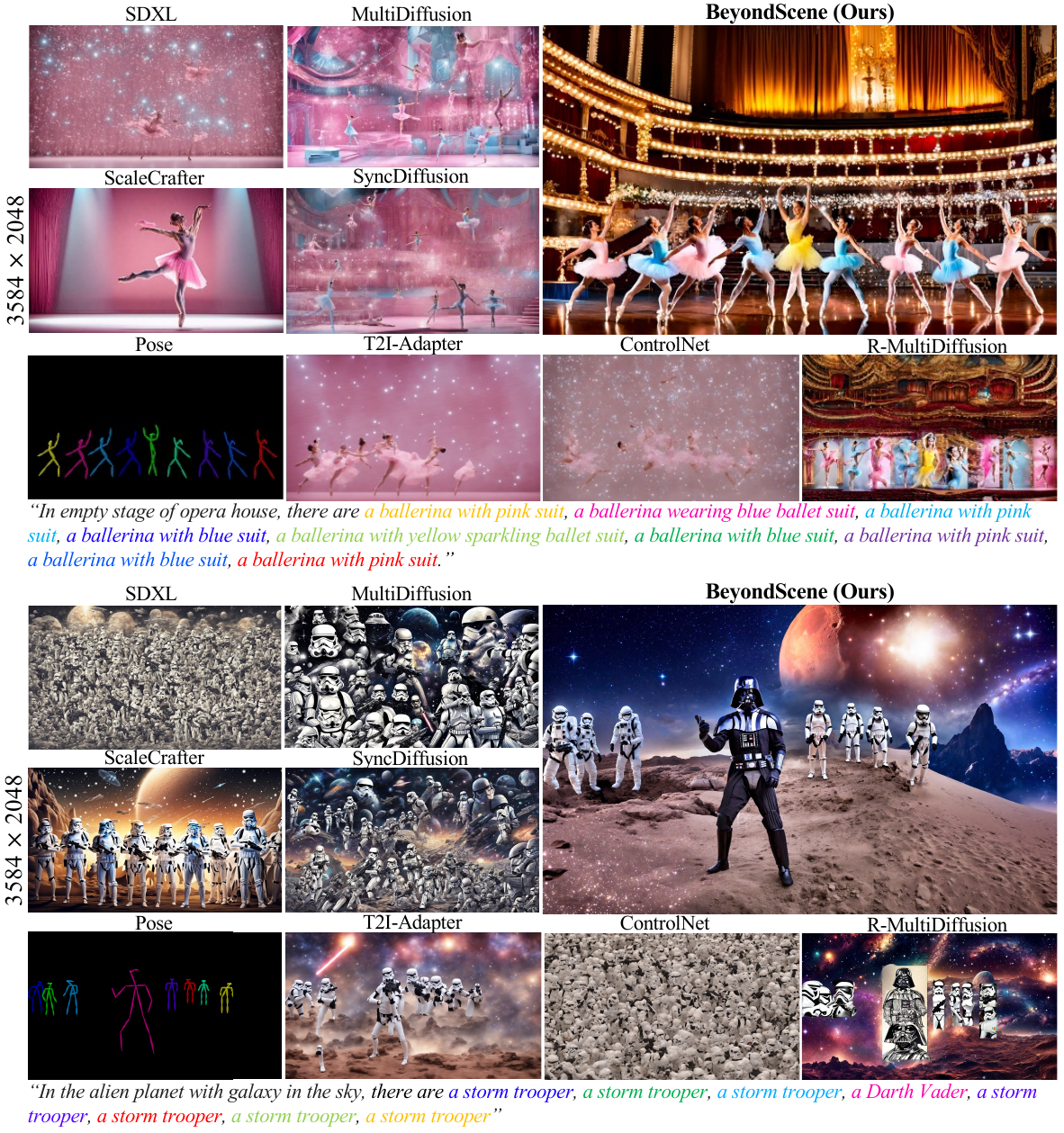}
   \vspace{-1em}
    \caption{Qualitative comparison for generating high-resolution human scenes (3584$\times$2048). While existing approaches like T2I-Direct (SDXL~\cite{podell2023sdxl}), T2I-Large (MultiDiffusion~\cite{bar2023multidiffusion}, SyncDiffusion~\cite{lee2023syncdiffusion}, and ScaleCrafter~\cite{he2023scalecrafter}), and Visual+T2I (ControlNet~\cite{podell2023sdxl, controlnet23}, T2IAdapter~\cite{podell2023sdxl,t2i23}, and R-MultiDiffusion~\cite{bar2023multidiffusion}) models struggle with artifacts, our method achieves superior results by producing images with minimal artifacts, strong text-image correspondence, and a natural look. The color in each description represents the description for each instance that has the same color in the pose map.}        
    \label{fig_quali_custom2}
   \vspace{-2em}
\end{figure}

\begin{figure}[!t]
    \centering
    \includegraphics[width=0.99\textwidth]{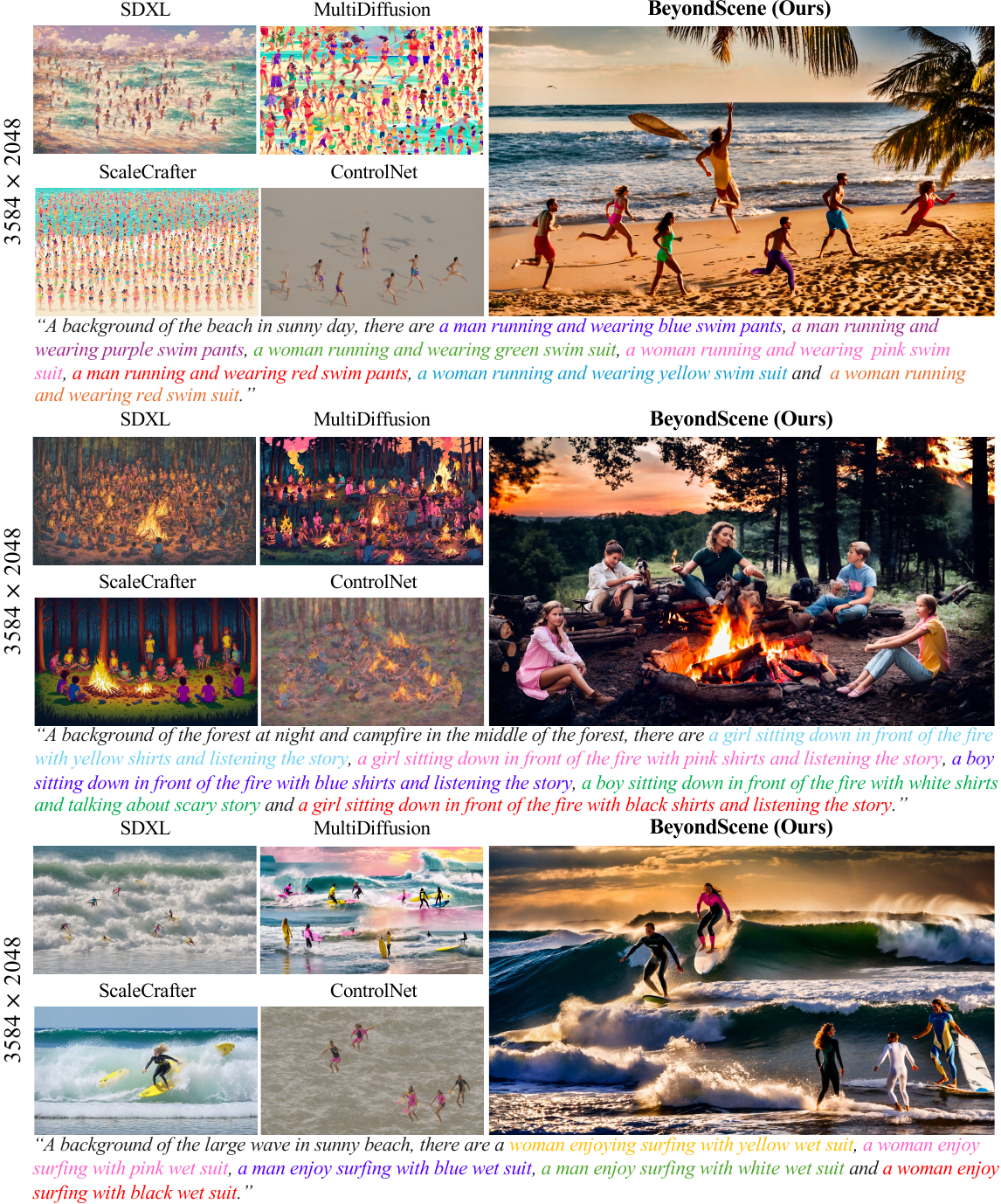}
  % \vspace{-1em}
    \caption{Additional custom examples of large scene synthesis (3584$\times$2048) for qualitative comparisons. Compared to existing approaches like SDXL~\cite{podell2023sdxl}, MultiDiffusion~\cite{bar2023multidiffusion},  ScaleCrafter~\cite{he2023scalecrafter}, and ControlNet~\cite{podell2023sdxl, controlnet23}, our method achieves minimal artifacts, strong text-image correspondence , and high global and human naturalness.}  
    \label{fig_quali_custom_3584}
  % \vspace{-1em}
\end{figure}

\begin{figure}[!t]
    \centering
    \includegraphics[width=0.99\textwidth]{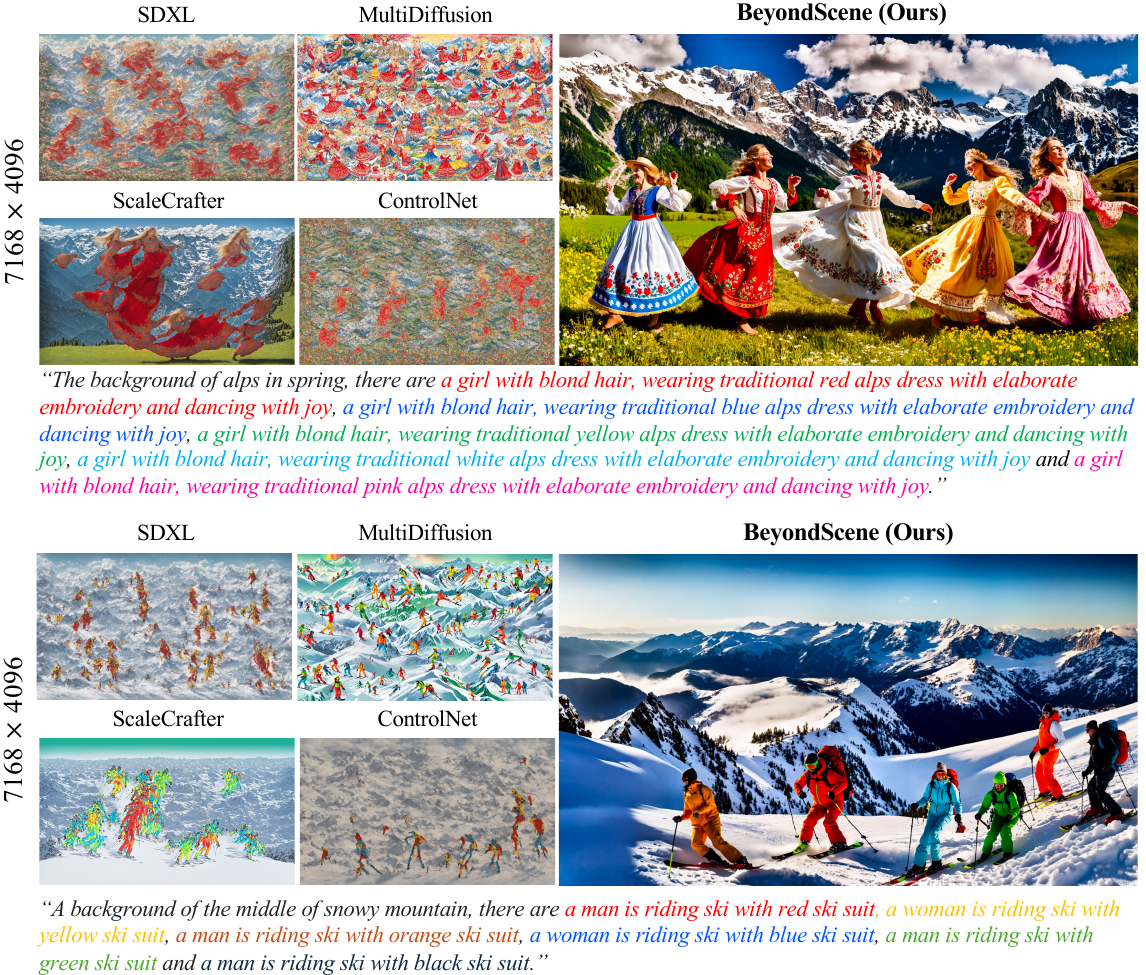}
  % \vspace{-1em}
    \caption{Additional custom examples of large scene synthesis (7186$\times$4096) for qualitative comparisons. Compared to existing approaches like SDXL~\cite{podell2023sdxl}, MultiDiffusion~\cite{bar2023multidiffusion}, ScaleCrafter~\cite{he2023scalecrafter}, and ControlNet~\cite{podell2023sdxl, controlnet23}, our method generates images that perfectly capture the essence of the text, appearing as natural as real-world scenes with realistic human depictions.}  
    \label{fig_quali_custom_7186}
  % \vspace{-1em}
\end{figure}

\begin{figure}[!t]
    \centering
    \includegraphics[width=0.99\textwidth]{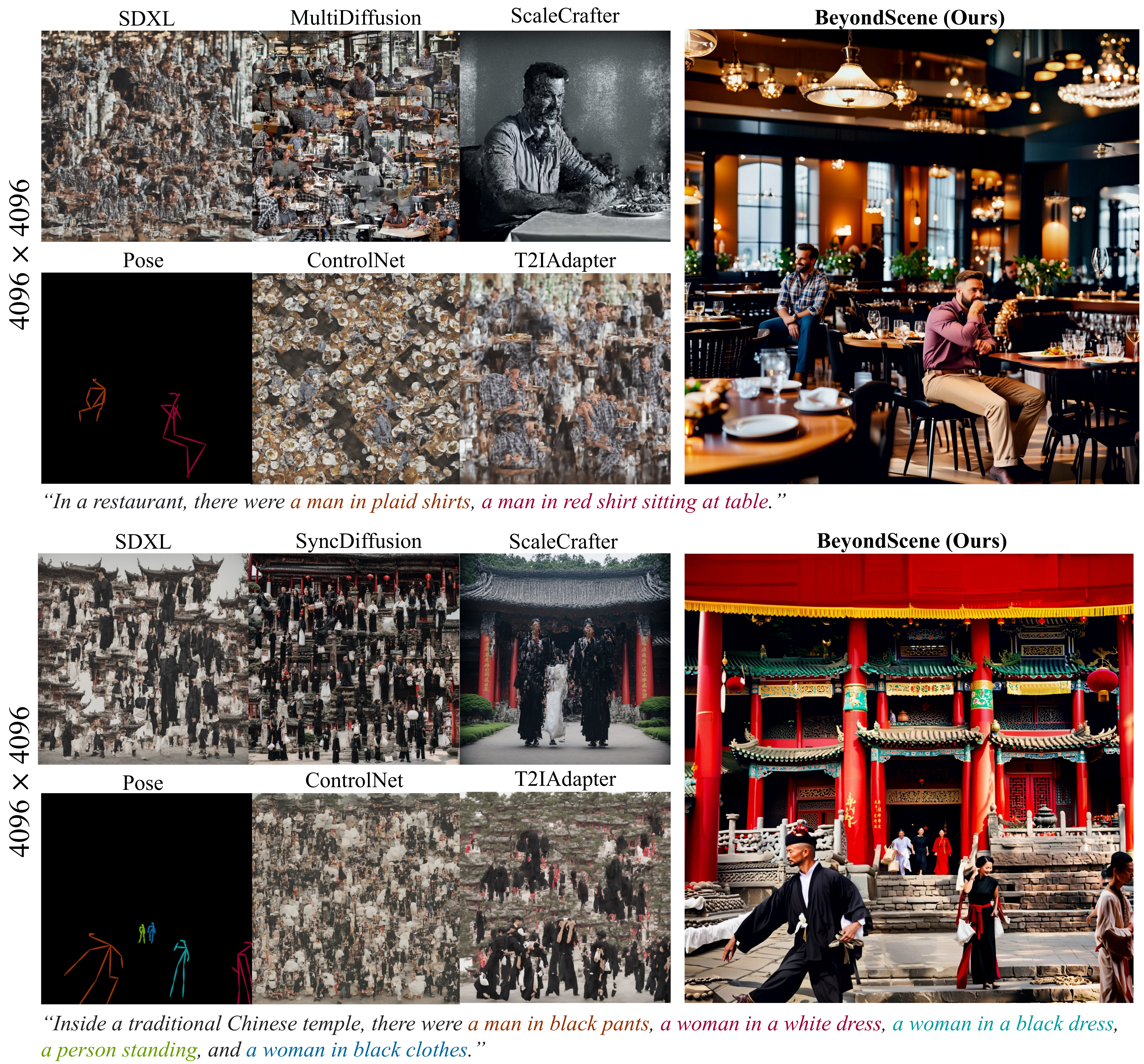}
   \vspace{-1em}
    \caption{Examples of large scene synthesis (4096$\times$4096) on the poses and text obtained from CrowdCaption~\cite{wang2022happens} images. 
    All baselines including direct high-resolution inference (SDXL~\cite{podell2023sdxl}), T2I models for large scenes (MultiDiffusion~\cite{bar2023multidiffusion}, SyncDiffusion~\cite{lee2023syncdiffusion}, ScaleCrafter~\cite{he2023scalecrafter}), Visual prior-guided T2I models (ControlNet~\cite{controlnet23}, T2IAdapter~\cite{t2i23}) produce duplicated objects and artifacts in human anatomy, while our method succeeded in generation of high-resolution image with high text-image correspondence. Each color in the description corresponds to instances sharing the same color in the pose map.}        
    \label{fig_quali_crowdcaption}
\vspace{-2em}

\end{figure}

\section{Experiments}

\subsection{Experimental Settings}
We comprehensively evaluate our method using both qualitative and quantitative metrics. To ensure a fair comparison, all compared models are implemented using Stable Diffusion XL (SDXL)~\cite{podell2023sdxl}-based architectures.  We utilize SDXL-ControlNet-Openpose~\cite{controlnet23, sdxl_controlnet_openpose}, SDXL-inpainting~\cite{sdxl_inpainting}, and Lang-Segment-Anything~\cite{lang_sam} for our pose-guided T2I diffusion, inpainting diffusion model, and segmentation models.
The training resolution is set to 1024$\times$1024, while inference resolutions are varied across 2048$\times$2048 (4$\times$ 1:1), 3584$\times$2048 (7$\times$ 7:4), 4096$\times$2048 (8$\times$ 2:1),  4096$\times$4096 (16$\times$ 1:1), and 8192$\times$8192 (64$\times$ 1:1).
\subsubsection{Testing data} Existing T2I generation datasets often lack large-scale scenes featuring multiple people with detailed descriptions for each individual. To address this limitation, we leverage the CrowdCaption dataset~\cite{wang2022happens} as our primary test set. This dataset features images containing large crowds of people, allowing us to evaluate the effectiveness of our method in generating complex scenes with numerous individuals.
For each image in the CrowdCaption dataset, we obtain descriptions and individual poses using GRIT~\cite{wu2022grit} and ViTPose~\cite{xu2022vitpose}. Additionally, we obtain a global caption generated using GPT4~\cite{achiam2023gpt}. 
We filter the CrowdCaption images based on two criteria: number of humans (max 8 to accommodate token limits of baselines) and aspect ratio (close to 1:1 and 2:1), which results in a collection of 100 images for each aspect ratio (1:1 and 2:1). Additionally, to showcase the versatility of our method beyond the CrowdCaption dataset, we incorporate custom examples for qualitative comparisons.
\subsubsection{Evaluation metrics} For an evaluation of \textit{text-image correspondence}, we adopt a multimodal large language model~\cite{achiam2023gpt, liu2024visual, liu2023llava}-based text-image correspondence metric, as proposed in VIEScore~\cite{ku2023viescore}, alongside the global CLIP~\cite{clip} score that shows its limitations in producing reliable scores for complex scenes with detailed descriptions due to limitations in input image size and the available number of input tokens. This MLLM-based metric, powered by GPT4~\cite{achiam2023gpt}, assigns a score ranging from 0 to 10, where 0 indicates no correspondence between the generated image and the prompt and 10 indicates perfect alignment. This metric has been demonstrably well-correlated with human judgments and provides explanations for the assigned scores~\cite{ku2023viescore}.
To assess the \textit{naturalness} of the generated images, we adapt the existing MLLM-based naturalness metric to focus on human-centric scenarios, leading to two separate scores: \textit{Global naturalness}  scores from 0 to 10, with 0 indicating an unnatural or unrealistic image and 10 indicating a completely natural image. This score considers factors such as overall inconsistencies, unrealistic physics, disconnectivity. \textit{Human naturalness}  scores from 0 to 10, with 0 indicating unnatural human anatomy, and 10 indicating the absence of any artifacts in the human anatomy.
Additional details on implementation and evaluation are provided in the supplementary material.

\begin{table}[!t]\centering
\caption{ Quantitative comparison of our method with various approaches, including direct high-resolution inference (T2I-Direct)~\cite{podell2023sdxl, yang2024mastering}, T2I models designed for large scenes (T2I-Large)~\cite{bar2023multidiffusion, lee2023syncdiffusion,he2023scalecrafter}, and Visual prior-guided T2I models (Visual+T2I)\cite{controlnet23, t2i23, bar2023multidiffusion}. }\label{tab_quant_main}
\scriptsize
 \begin{adjustbox}{width=0.8\linewidth}
\begin{tabular}{ccccccc}\toprule
\multirow{2}{*}{Model types} &\multirow{2}{*}{Models} &\multirow{2}{*}{Global CLIP} &\multicolumn{3}{c}{MLLM (GPT4)-based } \\\cmidrule{4-6}
& & &\makecell{Text-image\\correspondence} &\makecell{Global\\naturalness} &\makecell{Human\\naturalness} \\\midrule
\multicolumn{2}{c}{} &\multicolumn{4}{c}{2048$\times$2048 (4$\times$ 1:1)} \\\midrule
\multirow{2}{*}{T2I-Direct} &SDXL &0.324 &2.141 &1.838 &1.417 \\
&RPG &0.261 &1.963 &2.147 &2.785 \\\midrule
\multirow{3}{*}{T2I-Large} &MultiDiffusion &\textbf{0.347} &3.855 &3.381 &2.466 \\
&SyncDiffusion &\textbf{0.347} &3.655 &3.429 &2.607 \\
&ScaleCrafter &0.345 &6.081 &5.667 &5.062 \\\midrule
\multirow{3}{*}{Visual+T2I} &ControlNet &0.298 &1.790 &1.466 &1.117 \\
&T2IAdapter &0.328 &3.094 &2.660 &1.728 \\
&R-MultiDiffusion &0.311 &3.636 &1.711 &1.597 \\
&\textbf{BeyondScene} &0.305 &\textbf{7.041} &\textbf{6.535} &\textbf{6.114} \\\midrule
\multicolumn{2}{c}{} &\multicolumn{4}{c}{4096$\times$2048 (8$\times$ 2:1)} \\\midrule
\multirow{2}{*}{T2I-Direct} &SDXL &0.301 &1.061 &0.771 &0.607 \\
&RPG &OOM &OOM &OOM &OOM \\\midrule
\multirow{3}{*}{T2I-Large} &MultiDiffusion &0.338 &2.136 &2.228 &1.329 \\
&SyncDiffusion &\textbf{0.342} &3.128 &2.665 &1.591 \\
&ScaleCrafter &0.296 &3.552 &3.569 &3.359 \\\midrule
\multirow{3}{*}{Visual+T2I} &ControlNet &0.278 &1.526 &1.016 &0.657 \\
&T2IAdapter &0.282 &1.109 &1.010 &0.568 \\
&R-MultiDiffusion &0.306 &3.466 &1.422 &1.201 \\
&\textbf{BeyondScene} &0.304 &\textbf{7.118} &\textbf{6.612} &\textbf{6.375} \\\midrule
\multicolumn{2}{c}{} &\multicolumn{4}{c}{4096$\times$4096 (16$\times$ 1:1)} \\\midrule
\multirow{2}{*}{T2I-Direct} &SDXL &0.277 &0.329 &0.411 &0.118 \\
&RPG &OOM &OOM &OOM &OOM \\\midrule
\multirow{3}{*}{T2I-Large} &MultiDiffusion &\textbf{0.319} &1.305 &2.499 &1.591 \\
&SyncDiffusion &0.310 &0.912 &2.309 &1.359 \\
&ScaleCrafter &0.312 &2.962 &3.048 &2.442 \\\midrule
\multirow{3}{*}{Visual+T2I} &ControlNet &0.244 &0.811 &1.309 &0.729 \\
&T2IAdapter &0.269 &0.491 &0.719 &0.352 \\
&R-MultiDiffusion &0.281 &1.792 &0.860 &0.708 \\
&\textbf{BeyondScene} &0.301 &\textbf{6.801} &\textbf{6.074} &\textbf{5.627} \\
\bottomrule
\end{tabular}
 \end{adjustbox}
 % \vspace{-1em}

\end{table}

\begin{table}[!t]\centering
\caption{Quantitative comparison by scene complexity. Our method shows robust performance  (averaged across resolutions: 2048$\times$2048, 4096$\times$2048, 4096$\times$4096) even with increasing the number of humans, highlighting its ability to handle complex scenes.}\label{tab_quant_num_human}
\scriptsize
\begin{adjustbox}{width=0.75\linewidth}
\begin{tabular}{cccccccc}\toprule
\multirow{3}{*}{Models} &\multicolumn{3}{c}{2$\sim$4 humans} &\multicolumn{3}{c}{5$\sim$8 humans} \\\cmidrule(lr){2-4}\cmidrule(lr){5-7}
&\multicolumn{3}{c}{MLLM (GPT4)-based } &\multicolumn{3}{c}{MLLM (GPT4)-based } \\\cmidrule{2-7}
&\makecell{Text-image\\corr.}  &\makecell{Global\\nat.} &\makecell{Human\\nat.} &\makecell{Text-image\\corr.}  &\makecell{Global\\nat.} &\makecell{Human\\nat.}\\\midrule
SDXL &1.192 &1.039 &0.845 &1.161 &0.974 &0.853 \\
RPG &3.757 &3.138 &3.013 &0.169 &1.156 &2.558 \\
MultiDiffusion &2.142 &2.392 &1.692 &2.722 &3.013 &1.911 \\
SyncDiffusion &2.277 &2.428 &1.692 &2.853 &3.175 &2.013 \\
ScaleCrafter &4.55 &4.437 &4.064 &3.846 &3.752 &3.177 \\
ControlNet &1.172 &1.194 &0.810 &1.579 &1.333 &0.858 \\
T2IAdapter &1.521 &1.449 &0.964 &1.608 &1.486 &0.801 \\
R-MultiDIffusion &3.27 &1.286 &1.329 &2.659 &1.376 &1.007 \\
\textbf{BeyondScene} &\textbf{7.308} &\textbf{6.496} &\textbf{6.376} &\textbf{6.666} &\textbf{6.319} &\textbf{5.701} \\
\bottomrule
\end{tabular}
\end{adjustbox}
\end{table}

\setlength\intextsep{0.0em}
\setlength\columnsep{0.5em}
\setlength{\tabcolsep}{0.05em} % for the horizontal padding
\begin{table}[!t]\centering
\caption{User study result. Users consistently preferred our method over baselines, with MLLM scores closely aligning with human choices. However, the global CLIP score diverged, suggesting potential limitations. }\label{tab_quant_num_human}
\scriptsize
\begin{adjustbox}{width=0.48\linewidth}
\scriptsize
{\begin{tabular}{ccccc}\toprule
\multirow{2}{*}{Models} &\multicolumn{3}{c}{Preference of Ours ↑} \\\cmidrule{2-4}
&\makecell{Text-image\\corr.} &\makecell{Global\\nat.}  &\makecell{Human\\nat.} \\\midrule
vs ScaleCrafter &0.795 &0.650 &0.806 \\
vs MultiDiffusion &0.727 &0.658 &0.822 \\
vs ControlNet &0.870 &0.694 &0.897 \\
\bottomrule
\end{tabular}
}
\end{adjustbox}
\vspace{-2em}

\end{table}

\subsection{Result}
Our method is compared against three groups of existing SOTA approaches: 1) Direct high-resolution inference using T2I models (T2I-Direct) that directly perform high-resolution inference of  SDXL~\cite{podell2023sdxl} and RPG~\cite{yang2024mastering}  over their training image size,  2) T2I diffusion models for large scenes (T2I-Large) that encompass joint diffusion-based methods (MultiDiffusion~\cite{bar2023multidiffusion}, SyncDiffusion~\cite{lee2023syncdiffusion}), and ScaleCrafter~\cite{he2023scalecrafter}. ) Visual prior-guided T2I generation models (Visual+T2I) that focuses on models including SDXL-ControlNet-Openpose~\cite{podell2023sdxl, controlnet23}, SDXL-T2IAdapter~\cite{podell2023sdxl,t2i23}, and R-MultiDiffusion~\cite{bar2023multidiffusion} that leverage visual priors for T2I generation as our method. We provide more results in the supplementary material.

\subsubsection{Qualitative comparison} As represented in Fig.~\ref{fig_quali_custom1},~\ref{fig_quali_custom2},~\ref{fig_quali_custom_3584},~\ref{fig_quali_custom_7186} and~\ref{fig_quali_crowdcaption}, baselines introduce noticeable instance duplication, artifacts in anatomy, facial features, making the scene unnatural rather than reflecting real-world physics. Also, they struggle to generate complex scenes corresponding to the detailed descriptions. Conversely, our method generates images with minimal artifacts, demonstrating superior text-image correspondence, global and human naturalness, particularly in human-centric scenes.

\subsubsection{Quantitative comparison} Table~\ref{tab_quant_main} summarizes the performance across the resolution settings of  2048$\times$2048 (4$\times$ 1:1), 4096$\times$2048 (8$\times$ 2:1), and 4096$\times$4096 (16$\times$ 1:1). Our method consistently outperforms baselines on all metrics except the global CLIP score, which shows inconsistencies across methods. This suggests that our approach effectively enhances the original generation capabilities of the pretrained diffusion model for high-resolution image generation, achieving superior text-image correspondence, naturalness.

Table~\ref{tab_quant_num_human} further analyzes the performance of different methods based on the number of people depicted in the scene. Our approach demonstrates robustness, effectively managing complexity as people count rises.

\begin{figure}[!t]
    \centering
    \includegraphics[width=0.99\textwidth]{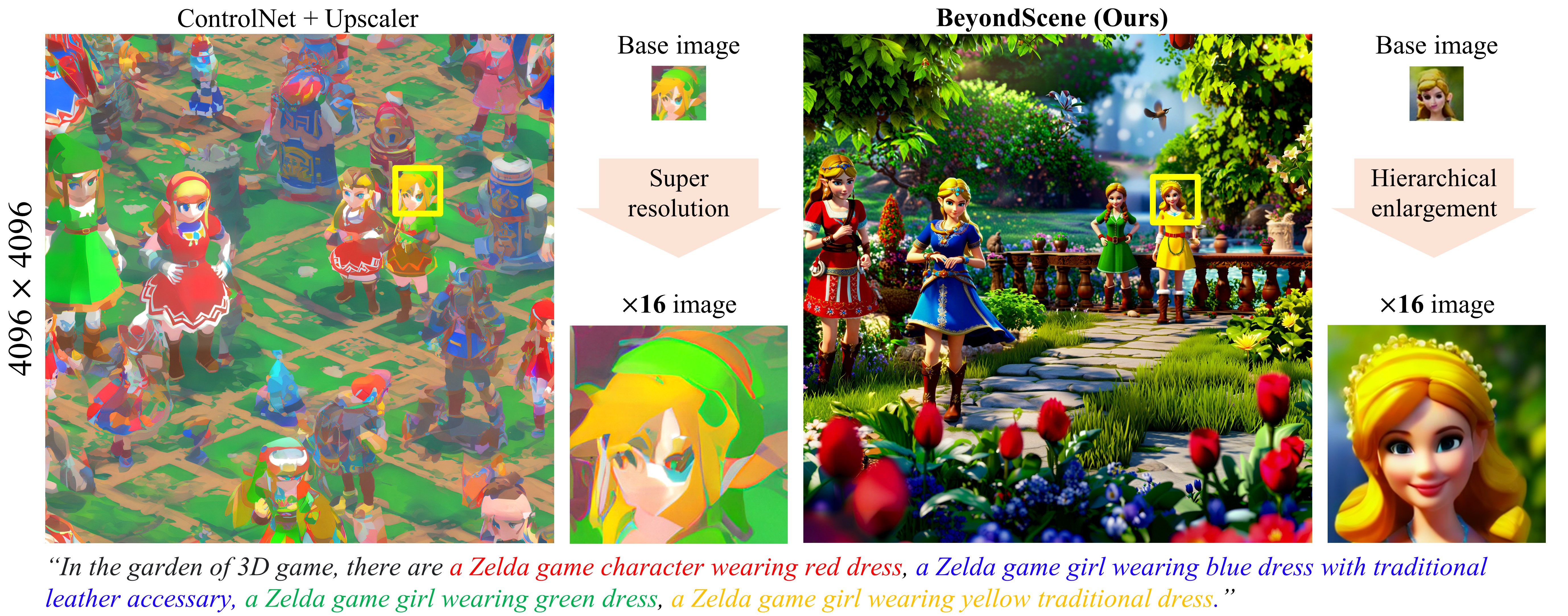}
  % \vspace{-1em}
    \caption{Comparison with SDXL-ControlNet-Openpose~\cite{controlnet23, sdxl_controlnet_openpose} combined with diffusion-based super-resolution (SD-Upscaler~\cite{sd_upscaler}) for high-resolution human-centric scene generation (4096$\times$4096). BeyondScene demonstrate superior text-image correspondence with more details. 
 }  
    \label{fig_sr}
  % \vspace{-1em}
      \centering
    \includegraphics[width=0.99\textwidth]{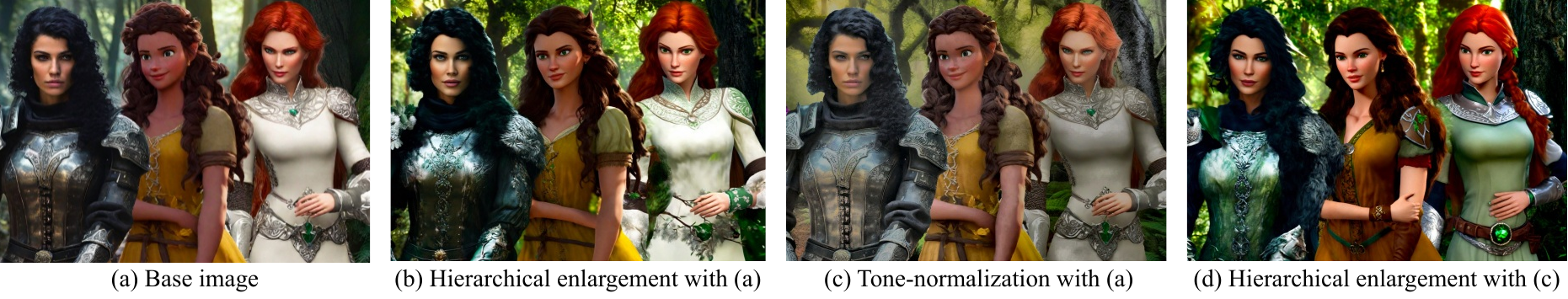}
   \vspace{-1em}
    \caption{Qualitative results on effectiveness of tone normalization. \textbf{(a)} The generated base image looks unnatural because the style and lighting vary between each instance. \textbf{(b)} Hierarchically enlarged image with (a) suffers from the same issues of (a). \textbf{(c)} Tone normalization with (a). \textbf{(d)} Hierarchically enlarged image with (c) exhibits uniformity in style and lighting, blending naturally into the background.}        
    \label{fig:ablation_tone}
    \label{fig:ablation_freq}
\end{figure}

\subsubsection{User study}  We conducted a user study to evaluate the faithfulness and naturalness of images generated by three methods (one from each group) and our method. 
Participants compared large-scene images based on the same criteria used in the MLLM metrics: text-image correspondence, global naturalness, and human naturalness. 
A total of 12,120 responses were collected from 101 participants.
The result shows that our method was consistently preferred over baselines, with a larger margin for all attributes. Notably, MLLM scores demonstrate strong alignment with human scores, supporting their validity. However, the global CLIP score appears misaligned, suggesting its limitations in capturing human judgment.
We provide the additional details in the supplementary material.

\subsubsection{Beyond 8K ultra-high-resolution image}
As displayed in Fig.~\ref{fig_quali_8k}, BeyondScene excels in generating unparalleled resolution, producing images at an impressive 8192$\times$8192 dimensions. This achievement surpasses the training resolution of SDXL by a remarkable factor of 64$\times$. It surpasses the standard 8K resolution (7680$\times$4320), demonstrating the model's capability to produce images with exceptional detail at even higher resolutions.

\subsubsection{Comparison with super resolution method} \label{sec:sr}
We compare BeyondScene's ability to generate high-resolution human-centric scenes with SDXL-ControlNet-Openpose~\cite{controlnet23, sdxl_controlnet_openpose} combined with diffusion-based super-resolution model (SD-Upscaler~\cite{sd_upscaler}). ControlNet generates a 1024$\times$1024 image conditioned on the same pose and text as our method, and then SD-Upscaler enlarges it to 4096$\times$4096. As shown in Fig.~\ref{fig_sr}, ControlNet+SD-SR struggles to generate images that faithfully correspond to the text description, missing key attributes. In contrast, BeyondScene's detailed base image generation enables it to produce scenes that closely match the text. Additionally, our instance-aware hierarchical enlargement progressively refines the scene elements from the base image in a clear and detailed way, whereas SD-SR simply increases the resolution with minimal improvement.

\begin{figure}[!t]
    \centering
    \includegraphics[width=0.99\textwidth]{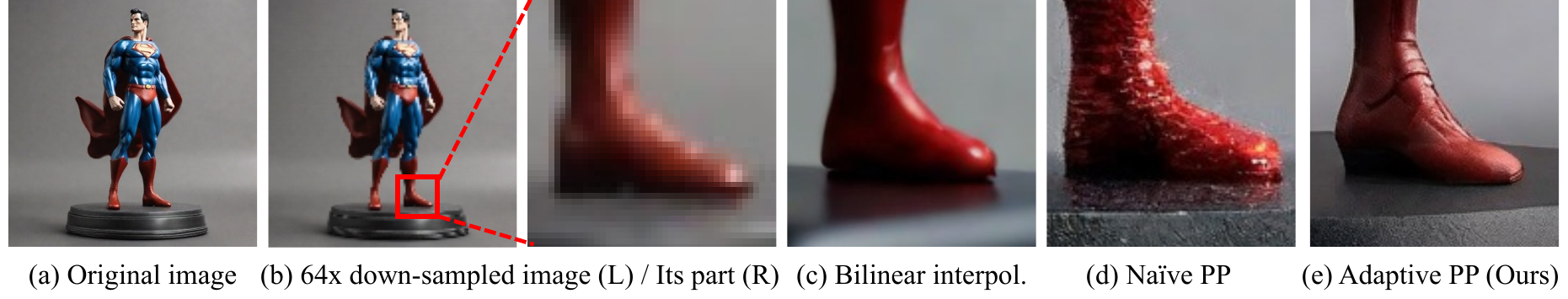}
   \vspace{-1em}
    \caption{Forward-reverse diffusion process with interpolation methods. \textbf{(a)} SDXL generates an image with the prompt \textit{``figurine of Superman''}. \textbf{(b)} The image is then $\times 64$ down-sampled (Left) with a close-up view (Right). \textbf{(c)} Upsampled image with bilinear interpolation and forward-reverse diffusion exhibits blurry details. \textbf{(d)} Naive pixel perturbation (PP) yields sharper details but introduces flickering artifacts. \textbf{(e)} Our Canny map-based adaptive PP produces clear and highly detailed image.}        
    \label{fig:ablation_freq}
        \centering
    \includegraphics[width=0.9\textwidth]{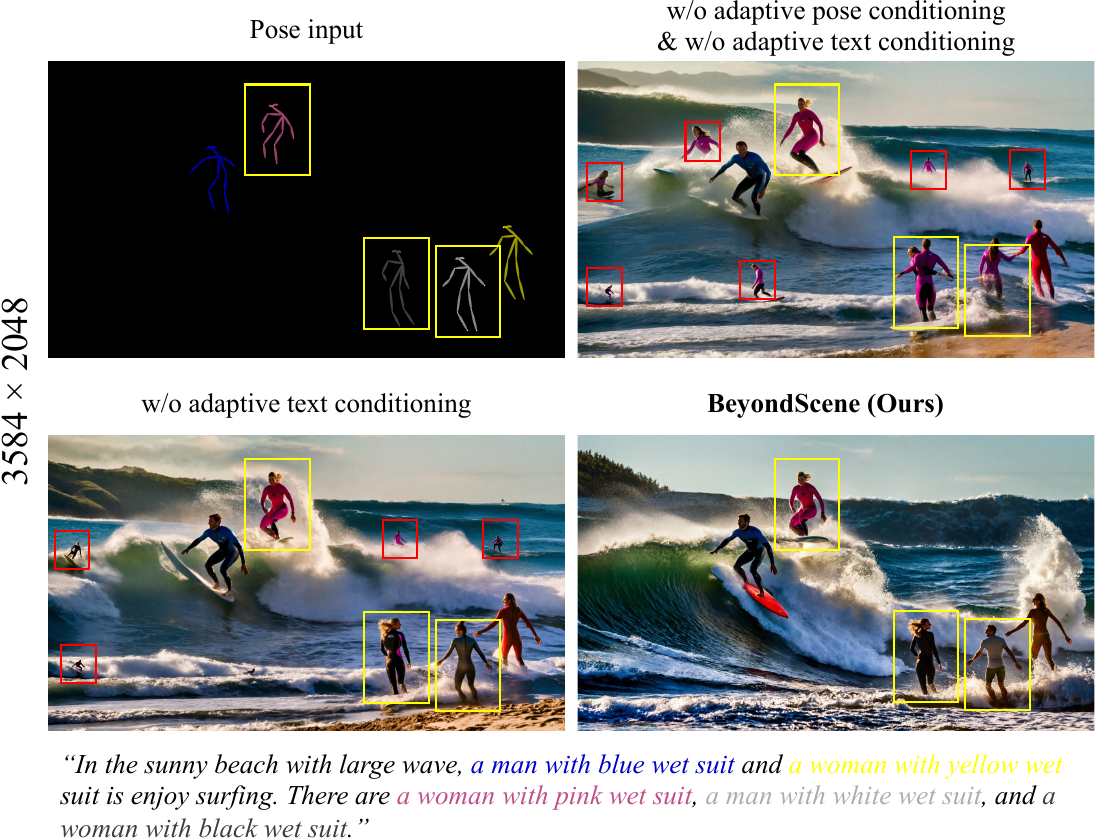}
  % \vspace{-1em}
    \caption{Effectiveness of adaptive conditioning.  Omitting either adaptive pose or text conditioning leads to issues like inconsistent poses, bad anatomy (yellow boxes), unwanted object duplication (red boxes), and mixed descriptions while BeyondScene with both mechanisms achieves maintains pose consistency, avoids  duplication, and ensures strong text-image correspondence. Each color in the description corresponds to instances sharing the same color in the pose map.}  
    \label{fig_ablation_adaptive_cond}
  % \vspace{-1em}
\end{figure}

% \begin{figure}[!t]
%     \centering
%     \includegraphics[width=0.99\textwidth]{figures/tone_norm.pdf}
%    \vspace{-1em}
%     \caption{Qualitative results on effectiveness of tone normalization. \textbf{(a)} The generated base image looks unnatural because the style and lighting vary between each instance. \textbf{(b)} Hierarchically enlarged image with (a) suffers from the same issues of (a). \textbf{(c)} Tone normalization with (a). \textbf{(d)} Hierarchically enlarged image with (c) exhibits uniformity in style and lighting, blending naturally into the background.}        
%     \label{fig:ablation_tone}
%     \centering
%     \includegraphics[width=0.99\textwidth]{figures/high_freq_ablation.pdf}
%    \vspace{-1em}
%     \caption{Forward-reverse diffusion process with interpolation methods. \textbf{(a)} SDXL generates an image with the prompt \textit{``figurine of Superman''}. \textbf{(b)} The image is then $\times 64$ down-sampled (Left) with a close-up view (Right). \textbf{(c)} Upsampled image with bilinear interpolation and forward-reverse diffusion exhibits blurry details. \textbf{(d)} Naive pixel perturbation (PP) yields sharper details but introduces flickering artifacts. \textbf{(e)} Our Canny map-based adaptive PP produces clear and highly detailed image.}        
%     \label{fig:ablation_freq}
%  %   \vspace{-1em}
% \end{figure}

\begin{figure}[!t]
      \centering
    \includegraphics[width=0.99\textwidth]{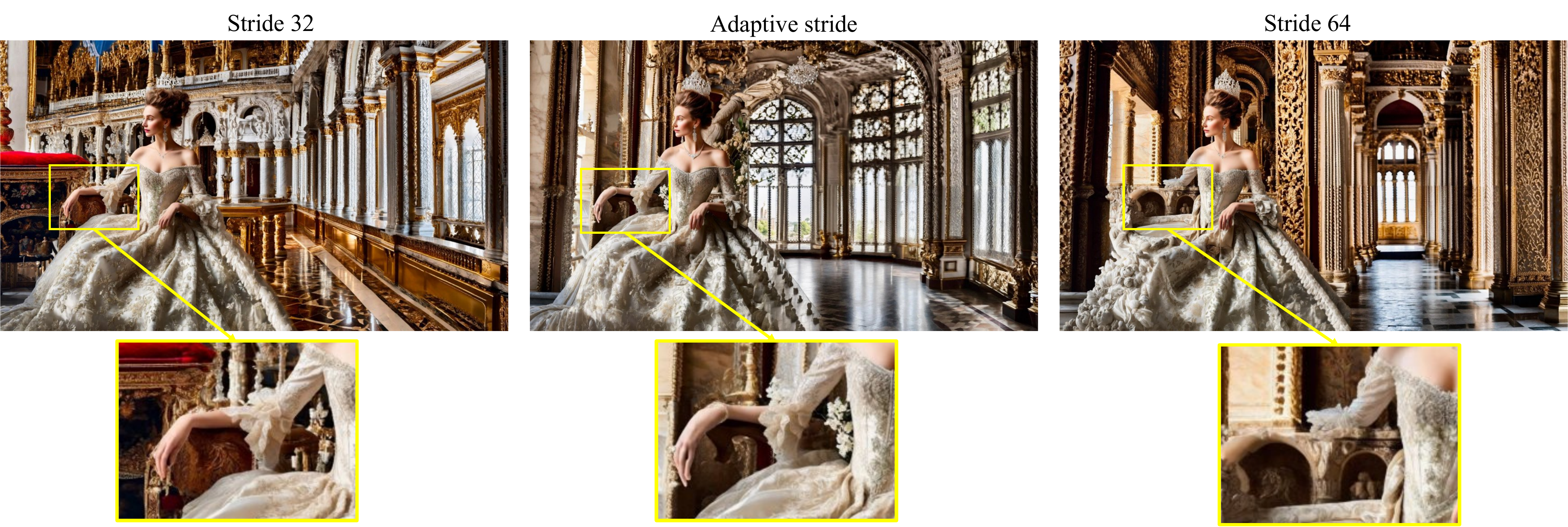}
  % \vspace{-1em}
    \caption{Effectiveness of adaptive stride in large scene ($3584\times 2048$). While the fixed stride of 32 in the leftmost image produces artifact-free results, it suffers from longer processing times. Conversely, the rightmost image with a stride of 64 generates the background efficiently, but introduces severe anatomical artifacts, like missing limbs. Our proposed adaptive stride method (middle) strikes a balance, achieving high-quality images with correct human anatomy while reducing processing time compared to the fixed stride of 32.}  
    \label{fig_ablation_adaptive_stride}
\end{figure}

\subsection{Ablation Study}
\subsubsection{Tone normalization}  As represented in Fig.~\ref{fig:ablation_tone}, removing tone normalization results in instances appearing lacking natural color and style harmonization, making the overall scene appear unnatural.

\subsubsection{High frequency-injected forward diffusion} As represented in ~\ref{fig:ablation_freq},  forward diffusion and reverse generation with naively up-sampled image lead to increased blurring in the image. By adopting the pixel perturbation, it make the result sharpened. However, it introduced flickering artifacts at the image borders. Applying adaptive pixel perturbation based on the Canny map significantly improves image quality and reduces flickering artifacts.

\subsubsection{Adaptive conditioning}
Fig.~\ref{fig_ablation_adaptive_cond} demonstrates the effectiveness of adaptive conditioning. Without both adaptive mechanisms, the model suffers. Omitting adaptive pose conditioning leads to inconsistent poses, bad anatomy, and decreased controllability compared to the desired input pose (yellow boxes). Conversely, excluding adaptive text conditioning, where each view receives the same full text including global and all instance descriptions, results in unwanted object duplication (red boxes) that ignores existing object scales, creating an awkward scene. Additionally, the descriptions of each instance become mixed, resulting in low text-image correspondence and decreased controllability. In contrast, BeyondScene with both adaptive pose and text conditioning achieves excellent text-image correspondence, maintains pose consistency, and avoids object duplication.

\subsubsection{Adaptive stride}
 As represented in Fig.~\ref{fig_ablation_adaptive_stride}, our proposed adaptive stride dynamically allocates smaller strides to critical regions within instances that require high fidelity, while utilizing larger strides in less demanding areas. This effectively reduces the overall computational cost while maintaining image quality.

\section{Conclusion}
We introduce BeyondScene, a novel framework capable of generating high-resolution human-centric scenes with exceptional text-image correspondence resolving artifacts beyond the token limits and beyond the training image size. BeyondScene overcomes limitations in generating complex human-centric scenes with detailed control. It achieves this through a novel two-stage process: first creating a low-resolution base focusing on key elements, then progressively refining it with high-resolution details. This method surpasses existing methods in generating high-fidelity scenes with control over human characteristics and number, while maintaining naturalness and reducing artifacts.

\section*{Acknowledgement}
This work was supported by Institute of Information \& communications Technology Planning \& Evaluation (IITP) grant funded by the Korea government(MSIT) [NO.2021-0-01343, Artificial Intelligence Graduate School Program (Seoul National University)], by the National Research Foundation of Korea(NRF) grant funded by the Korea government(MSIT) (No. NRF-2022R1A4A1030579, NRF-2022M3C1A309202211) and Creative-Pioneering Researchers Program through Seoul National University. Also, the authors acknowledged the financial support from the BK21 FOUR program of the Education and Research Program for Future ICT Pioneers, Seoul National University.

%\clearpage  % TODO REVIEW/FINAL: This \clearpage needs to be removed from both review and camera-ready versions.

\bibliographystyle{splncs04}
\bibliography{main}

\begin{thebibliography}{10}
\providecommand{\url}[1]{\texttt{#1}}
\providecommand{\urlprefix}{URL }
\providecommand{\doi}[1]{https://doi.org/#1}

\bibitem{lang_sam}
Language segment anything. \url{https://github.com/luca-medeiros/lang-segment-anything}

\bibitem{midjourney}
Midjourney. \url{https://www.midjourney.com}

\bibitem{sdxl_controlnet_openpose}
Sdxl-controlnet: Openpose (v2). \url{https://huggingface.co/thibaud/controlnet-openpose-sdxl-1.0}

\bibitem{sdxl_inpainting}
Sdxl inpainting 0.1. \url{https://huggingface.co/diffusers/stable-diffusion-xl-1.0-inpainting-0.1}

\bibitem{sd_upscaler}
Stable diffusion x4 upscaler. \url{https://huggingface.co/stabilityai/stable-diffusion-x4-upscaler}

\bibitem{achiam2023gpt}
Achiam, J., Adler, S., Agarwal, S., Ahmad, L., Akkaya, I., Aleman, F.L., Almeida, D., Altenschmidt, J., Altman, S., Anadkat, S., et~al.: Gpt-4 technical report. arXiv preprint arXiv:2303.08774  (2023)

\bibitem{bar2023multidiffusion}
Bar-Tal, O., Yariv, L., Lipman, Y., Dekel, T.: Multidiffusion: Fusing diffusion paths for controlled image generation. arXiv:2302.08113  (2023)

\bibitem{chen2023importance}
Chen, T.: On the importance of noise scheduling for diffusion models. arXiv preprint arXiv:2301.10972  (2023)

\bibitem{kpe22_vae}
Cheong, S.Y., Mustafa, A., Gilbert, A.: {KPE}: Keypoint pose encoding for transformer-based image generation. In: British Machine Vision Conference (BMVC) (2022)

\bibitem{ding2023patched}
Ding, Z., Zhang, M., Wu, J., Tu, Z.: Patched denoising diffusion models for high-resolution image synthesis. In: The Twelfth International Conference on Learning Representations (2023)

\bibitem{he2023scalecrafter}
He, Y., Yang, S., Chen, H., Cun, X., Xia, M., Zhang, Y., Wang, X., He, R., Chen, Q., Shan, Y.: Scalecrafter: Tuning-free higher-resolution visual generation with diffusion models. In: The Twelfth International Conference on Learning Representations (2023)

\bibitem{ho2020denoising}
Ho, J., Jain, A., Abbeel, P.: Denoising diffusion probabilistic models. arXiv:2006.11239  (2020)

\bibitem{hoogeboom2023simple}
Hoogeboom, E., Heek, J., Salimans, T.: simple diffusion: End-to-end diffusion for high resolution images. arXiv preprint arXiv:2301.11093  (2023)

\bibitem{ju2023humansd}
Ju, X., Zeng, A., Zhao, C., Wang, J., Zhang, L., Xu, Q.: Humansd: A native skeleton-guided diffusion model for human image generation. arXiv preprint arXiv:2304.04269  (2023)

\bibitem{ku2023viescore}
Ku, M., Jiang, D., Wei, C., Yue, X., Chen, W.: Viescore: Towards explainable metrics for conditional image synthesis evaluation. arXiv preprint arXiv:2312.14867  (2023)

\bibitem{lee2023syncdiffusion}
Lee, Y., Kim, K., Kim, H., Sung, M.: Syncdiffusion: Coherent montage via synchronized joint diffusions. arXiv:2306.05178  (2023)

\bibitem{gligen23}
Li, Y., Liu, H., Wu, Q., Mu, F., Yang, J., Gao, J., Li, C., Lee, Y.J.: {GLIGEN}: Open-set grounded text-to-image generation. arXiv preprint arXiv:2301.07093  (2023)

\bibitem{verbalpersonnets22}
Liu, D., Wu, L., Zheng, F., Liu, L., Wang, M.: Verbal-person nets: Pose-guided multi-granularity language-to-person generation. IEEE Transactions on Neural Networks and Learning Systems  (2022)

\bibitem{liu2024visual}
Liu, H., Li, C., Wu, Q., Lee, Y.J.: Visual instruction tuning. Advances in neural information processing systems  \textbf{36} (2024)

\bibitem{liu2022pseudo}
Liu, L., Ren, Y., Lin, Z., Zhao, Z.: Pseudo numerical methods for diffusion models on manifolds. arXiv:2202.09778  (2022)

\bibitem{liu2023llava}
Liu, S., Cheng, H., Liu, H., Zhang, H., Li, F., Ren, T., Zou, X., Yang, J., Su, H., Zhu, J., et~al.: Llava-plus: Learning to use tools for creating multimodal agents. arXiv preprint arXiv:2311.05437  (2023)

\bibitem{spgnet21_gan}
Lv, Z., Li, X., Li, X., Li, F., Lin, T., He, D., Zuo, W.: Learning semantic person image generation by region-adaptive normalization. In: Proceedings of the IEEE/CVF Conference on Computer Vision and Pattern Recognition (CVPR). pp. 10806--10815 (2021)

\bibitem{poseguided17_gan}
Ma, L., Jia, X., Sun, Q., Schiele, B., Tuytelaars, T., Van~Gool, L.: Pose guided person image generation. vol.~30 (2017)

\bibitem{mustgan21_gan}
Ma, T., Peng, B., Wang, W., Dong, J.: {MUST-GAN}: Multi-level statistics transfer for self-driven person image generation. In: Proceedings of the IEEE/CVF Conference on Computer Vision and Pattern Recognition (CVPR). pp. 13622--13631 (2021)

\bibitem{decomposed20_gan}
Men, Y., Mao, Y., Jiang, Y., Ma, W.Y., Lian, Z.: Controllable person image synthesis with attribute-decomposed {GAN}. In: Proceedings of the IEEE/CVF Conference on Computer Vision and Pattern Recognition (CVPR). pp. 5084--5093 (2020)

\bibitem{t2i23}
Mou, C., Wang, X., Xie, L., Zhang, J., Qi, Z., Shan, Y., Qie, X.: T2{I-A}dapter: Learning adapters to dig out more controllable ability for text-to-image diffusion models. arXiv preprint arXiv:2302.08453  (2023)

\bibitem{podell2023sdxl}
Podell, D., English, Z., Lacey, K., Blattmann, A., Dockhorn, T., M{\"u}ller, J., Penna, J., Rombach, R.: Sdxl: Improving latent diffusion models for high-resolution image synthesis. arXiv preprint arXiv:2307.01952  (2023)

\bibitem{clip}
Radford, A., Kim, J.W., Hallacy, C., Ramesh, A., Goh, G., Agarwal, S., Sastry, G., Askell, A., Mishkin, P., Clark, J., Krueger, G., Sutskever, I.: Learning transferable visual models from natural language supervision. In: International Conference on Machine Learning (ICML). pp. 8748--8763. PMLR (2021)

\bibitem{ramesh2022hierarchical}
Ramesh, A., Dhariwal, P., Nichol, A., Chu, C., Chen, M.: Hierarchical text-conditional image generation with clip latents. arXiv preprint arXiv:2204.06125  \textbf{1}(2), ~3 (2022)

\bibitem{nted22_gan}
Ren, Y., Fan, X., Li, G., Liu, S., Li, T.H.: Neural texture extraction and distribution for controllable person image synthesis. In: Proceedings of the IEEE/CVF Conference on Computer Vision and Pattern Recognition (CVPR). pp. 13535--13544 (2022)

\bibitem{transformation20_vae}
Ren, Y., Yu, X., Chen, J., Li, T.H., Li, G.: Deep image spatial transformation for person image generation. In: Proceedings of the IEEE/CVF Conference on Computer Vision and Pattern Recognition (CVPR). pp. 7690--7699 (2020)

\bibitem{rombach2022highresolution}
Rombach, R., Blattmann, A., Lorenz, D., Esser, P., Ommer, B.: High-resolution image synthesis with latent diffusion models. In: CVPR. pp. 10684--10695 (2022)

\bibitem{rombach2022high}
Rombach, R., Blattmann, A., Lorenz, D., Esser, P., Ommer, B.: High-resolution image synthesis with latent diffusion models. In: CVPR. pp. 10684--10695 (2022)

\bibitem{rombach2021highresolution}
Rombach, R., Blattmann, A., Lorenz, D., Esser, P., Ommer, B.: High-resolution image synthesis with latent diffusion models (2021)

\bibitem{tips22}
Roy, P., Ghosh, S., Bhattacharya, S., Pal, U., Blumenstein, M.: {TIPS}: Text-induced pose synthesis. In: European Conference on Computer Vision (ECCV). pp. 161--178. Springer (2022)

\bibitem{saharia2022photorealistic}
Saharia, C., Chan, W., Saxena, S., Li, L., Whang, J., Denton, E.L., Ghasemipour, K., Gontijo~Lopes, R., Karagol~Ayan, B., Salimans, T., et~al.: Photorealistic text-to-image diffusion models with deep language understanding. NeurIPS  \textbf{35},  36479--36494 (2022)

\bibitem{song2020denoising}
Song, J., Meng, C., Ermon, S.: Denoising diffusion implicit models. arXiv:2010.02502  (2020)

\bibitem{teng2023relay}
Teng, J., Zheng, W., Ding, M., Hong, W., Wangni, J., Yang, Z., Tang, J.: Relay diffusion: Unifying diffusion process across resolutions for image synthesis. arXiv preprint arXiv:2309.03350  (2023)

\bibitem{yolov8}
ultralytics: yolov8. \url{https://github.com/ultralytics/ultralytics} (2023)

\bibitem{wang2022happens}
Wang, L., Li, H., Hu, W., Zhang, X., Qiu, H., Meng, F., Wu, Q.: What happens in crowd scenes: A new dataset about crowd scenes for image captioning. IEEE Transactions on Multimedia  (2022)

\bibitem{wu2022grit}
Wu, J., Wang, J., Yang, Z., Gan, Z., Liu, Z., Yuan, J., Wang, L.: Grit: A generative region-to-text transformer for object understanding. arXiv:2212.00280  (2022)

\bibitem{xie2023difffit}
Xie, E., Yao, L., Shi, H., Liu, Z., Zhou, D., Liu, Z., Li, J., Li, Z.: Difffit: Unlocking transferability of large diffusion models via simple parameter-efficient fine-tuning. arXiv preprint arXiv:2304.06648  (2023)

\bibitem{sharedspace21}
Xu, X., Chen, Y.C., Tao, X., Jia, J.: Text-guided human image manipulation via image-text shared space. IEEE Transactions on Pattern Analysis and Machine Intelligence (PAMI)  \textbf{44}(10),  6486--6500 (2021)

\bibitem{xu2022vitpose}
Xu, Y., Zhang, J., Zhang, Q., Tao, D.: Vitpose: Simple vision transformer baselines for human pose estimation. NeurIPS  \textbf{35},  38571--38584 (2022)

\bibitem{ctnet21_gan}
Yang, F., Lin, G.: {CT-Net}: Complementary transferring network for garment transfer with arbitrary geometric changes. In: Proceedings of the IEEE/CVF Conference on Computer Vision and Pattern Recognition (CVPR). pp. 9899--9908 (2021)

\bibitem{yang2024mastering}
Yang, L., Yu, Z., Meng, C., Xu, M., Ermon, S., Cui, B.: Mastering text-to-image diffusion: Recaptioning, planning, and generating with multimodal llms. arXiv preprint arXiv:2401.11708  (2024)

\bibitem{finegrained21_vae}
Yang, L., Wang, P., Liu, C., Gao, Z., Ren, P., Zhang, X., Wang, S., Ma, S., Hua, X., Gao, W.: Towards fine-grained human pose transfer with detail replenishing network. IEEE Transactions on Image Processing  \textbf{30},  2422--2435 (2021)

\bibitem{zhang2022styleswin}
Zhang, B., Gu, S., Zhang, B., Bao, J., Chen, D., Wen, F., Wang, Y., Guo, B.: Styleswin: Transformer-based gan for high-resolution image generation. In: Proceedings of the IEEE/CVF conference on computer vision and pattern recognition. pp. 11304--11314 (2022)

\bibitem{pise21_gan}
Zhang, J., Li, K., Lai, Y.K., Yang, J.: {PISE}: Person image synthesis and editing with decoupled {GAN}. In: Proceedings of the IEEE/CVF Conference on Computer Vision and Pattern Recognition (CVPR). pp. 7982--7990 (2021)

\bibitem{controlnet23}
Zhang, L., Agrawala, M.: Adding conditional control to text-to-image diffusion models. arXiv preprint arXiv:2302.05543  (2023)

\bibitem{dptn22_gan}
Zhang, P., Yang, L., Lai, J.H., Xie, X.: Exploring dual-task correlation for pose guided person image generation. In: Proceedings of the IEEE/CVF Conference on Computer Vision and Pattern Recognition (CVPR). pp. 7713--7722 (2022)

\bibitem{zhang2023diffcollage}
Zhang, Q., Song, J., Huang, X., Chen, Y., Liu, M.Y.: Diffcollage: Parallel generation of large content with diffusion models. arXiv:2303.17076  (2023)

\bibitem{zheng2023any}
Zheng, Q., Guo, Y., Deng, J., Han, J., Li, Y., Xu, S., Xu, H.: Any-size-diffusion: Toward efficient text-driven synthesis for any-size hd images. arXiv preprint arXiv:2308.16582  (2023)

\end{thebibliography}

\clearpage
% \appendix

% \twocolumn[{%
% \renewcommand\twocolumn[1][]{#1}%

% \begin{center}
% \bigskip 
% \bigskip 
% \textbf{\Large DATID-3D: Diversity-Preserved Domain Adaptation \\ Using Text-to-Image Diffusion for 3D Generative Model \\ (Supplementary Material) \\}

% \bigskip 
% \bigskip 
% {\large  Gwanghyun Kim$^1$ \qquad Se Young Chun$^{1,2}$ \\
% $^1$Dept. of Electrical and Computer Engineering, $^2$INMC, Interdisciplinary Program in AI\\
% Seoul National University, Korea \\
% {\tt\small \{gwang.kim, sychun\}@snu.ac.kr}
% }
% \bigskip 
% \bigskip 
% \maketitle
 
% \end{center}%
% }]

\setcounter{equation}{0}
\setcounter{figure}{0}
\setcounter{table}{0}
\setcounter{page}{1}
\makeatletter
\renewcommand{\theequation}{S\arabic{equation}}
\renewcommand{\thefigure}{S\arabic{figure}}
\renewcommand{\thetable}{S\arabic{table}}

\appendix

\titlerunning{BeyondScene}
\authorrunning{Kim et al.}
%%%%%%%%% TITLE

\setcounter{equation}{0}
\setcounter{figure}{0}
\setcounter{table}{0}
\setcounter{page}{1}
\makeatletter
\renewcommand{\theequation}{S\arabic{equation}}
\renewcommand{\thefigure}{S\arabic{figure}}
\renewcommand{\thetable}{S\arabic{table}}
\renewcommand\thesection{S\arabic{section}} % S added to 

% \title{BeyondScene: Higher-Resolution Human-Centric Scene Generation With Pretrained Diffusion\\(Supplementary Material)}

% \author{Gwanghyun Kim\inst{1\star},
% Hayeon Kim\inst{1\star},
% Hoigi Seo\inst{1\star},
% Dong Un Kang\inst{1}\thanks{Authors contributed equally. \ $^\dagger$Corresponding author.}, \\
% Se Young Chun\inst{1,2}$^{\dagger}$}

% % TODO FINAL: Replace with an abbreviated list of authors.


% % TODO FINAL: Replace with your institution list.
% \institute{$^1$Dept. of Electrical and Computer Engineering, $^2$INMC \&  IPAI  \\
% Seoul National University, Republic of Korea \\
% \email{\{gwang.kim, qkrtnskfk23, seohoiki3215, khy5630,  sychun\}@snu.ac.kr}   } 

% \maketitle
%\thispagestyle{empty}
%%%%%%%%%%%
\begin{center}
\bigskip 
\bigskip 
\textbf{\Large Supplementary Material \\}
\bigskip 
\bigskip 
 
\end{center}%

%%%%%%%%%%%%%%%%%%%%%%%%%%%%%%%%%%%%%%%%%%%%%%%%%%%
\section{More Details on Our Proposed Method}

\subsection{Instance-Aware Hierarchical Enlargement}

Instance-aware hierarchical enlargement takes a low-resolution image and enlarges it in stages. First, it upsamples the image, then injects high-frequency details using our proposed \textit{high frequency-injected forward diffusion}. Subsequently, our \textit{adaptive joint diffusion} enables to generate a higher resolution image using adaptive conditioning and adaptive stride. 
Here, we provide more detailed procedures with the algorithms for deeper understanding.

\subsubsection{High frequency-injected forward diffusion}
 BeyondScene proposes a novel adaptive pixel perturbation method that leverages edge information for generating fine-detailed and sharp images. The blurred edge map is generated with a Canny edge map and subsequent Gaussian smoothing. This blurred edge map is then normalized and conditioned to create a probability map ($\mathcal{C}$) that guides the perturbation process. We strategically perturb pixels in an interpolated image $\mathcal{I}_{\textnormal{p}}$ based on $\mathcal{C}$, selecting replacement pixels from the original image to preserve high-frequency details. The perturbed image $\mathcal{I}_{\textnormal{p}}$ is encoded with the variational autoencoder (VAE) to yield $\V{z}_0$, and then added noise of timestep $T_b$ with forward process results in $\V{z}_{T_b}$. The complete algorithm is presented in Algorithm~\ref{algo:pp}. $\mathcal{I}$ denotes the image from previous stage which has the height of $H$ and width of $W$, $\mathcal{I}_{\textnormal{p}}$ refers to the interpolated image which has the dimension of $\mathbb{R}^{\alpha_{\textnormal{interp}}H \times \alpha_{\textnormal{interp}}W}$, $d_r$ denotes the maximum distance of pixel to replace, $\sigma$ is the standard deviation of the Gaussian kernel for smoothing, $\alpha_{\textnormal{interp}}$ is the ratio of image expansion, $p_{\textnormal{max}}$ denotes the maximum and $p_{\textnormal{base}}$ means the minimum probability of the probability map $\mathcal{C}$.

\begin{algorithm}[!h]
%\ContinuedFloat
\caption{High frequency-injected forward diffusion}

\label{algo:pp}
\textbf{Input:} $\mathcal{I} \in \mathbb{R}^{H \times W}$(low-resolution image generated from previous adaptive Joint process), $d_r \in \mathbb{Z}$(maximum pixel distance for the pixel perturbation), $\sigma$(standard deviation of the Gaussian kernel for Canny map blurring), $\alpha_{\textnormal{interp}}$(ratio of enlargement), $p_{\textnormal{max}}$(maximum probability of probability map $\mathcal{C}$), $p_{\textnormal{base}}$(minimum probability of probability map $\mathcal{C}$), $T_b$(timestep for forward process)

\textbf{Output:} $\V{z}_{T_b}$
\SetKwProg{Fn}{Function}{:}{}
\SetKwFunction{FPP}{AdaptivePixelPert}
\medskip

\Fn{\FPP{$\mathcal{I}$, $d_r$, $\sigma$, $\alpha_{\textnormal{interp}}$, $p_{\textnormal{max}}$, $p_{\textnormal{base}}$}} {
    \bluetext{\tcp{Generate probability map $\mathcal{C}$}}
    \SetKwFunction{FCANNY}{Canny}
    $\hat{C}$ $\leftarrow$ \FCANNY{$\mathcal{I}, k_{\textnormal{min}}, k_{\textnormal{max}}$} \\
    
    \SetKwFunction{FBLUR}{GaussianBlur}
    $\hat{C}_G \leftarrow$ \FBLUR($\hat{C}, \sigma$)\\
    $C \leftarrow (p_{\textnormal{max}} - p_{\textnormal{base}}) \cdot \hat{C}_G + p_{\textnormal{base}}$ \\
    \medskip
    \bluetext{\tcp{Upsample image and probability map}}
    $H^*, W^* \leftarrow \alpha_{\textnormal{interp}}\cdot H, 
    \alpha_{\textnormal{interp}}\cdot W$ \\
    \SetKwFunction{FINTER}{LanczosInterp}
    $\mathcal{I}_{\textnormal{p}} \leftarrow \FINTER(\mathcal{I}, H^*, W^*)$ \\
    $\mathcal{C} \leftarrow \FINTER( \hat{C}_G, H^*, W^*)$ \\
    
    \medskip
    \SetKwFunction{FRAND}{RANDINT}
    \SetKwFunction{FMIN}{MIN}
    \SetKwFunction{FMAX}{MAX}
    \bluetext{\tcp{Pixel perturbation-based on probability map}}
    \For{$h=0, \dots ,\ H^*-1$}{ 
        \For{$w=0, \dots ,\ W^*-1$}{
            \If{$\epsilon_{h, w} \sim \mathcal{U}(0, 1)$ \textgreater \ $\mathcal{C}_{h, w}$}{
                $h_{\textnormal{rand}} \leftarrow$ \FMIN{\FMAX{$h / \alpha_{\textnormal{interp}}$ + \FRAND{$-d_r, d_r$}, 0}, H} \\
                $w_{\textnormal{rand}} \leftarrow$ \FMIN{\FMAX{$w / \alpha_{\textnormal{interp}}$ + \FRAND{$-d_r, d_r$}, 0}, W} \\
                $\mathcal{I}_{\textnormal{p}_{h, w}} \leftarrow \mathcal{I}_{h_{\textnormal{rand}}, w_{\textnormal{rand}}}$
            }
            % \Else{
            % pass
            % }
        }
    }
    \Return{$\mathcal{I}_{\textnormal{p}}$}
}
\medskip

\bluetext{\tcp{High frequency-injected forward diffusion}}
$\mathcal{I}_{p} \leftarrow$ \FPP{$\mathcal{I}$, $d_r$, $\sigma$, $\alpha_{\textnormal{interp}}$, $p_{\textnormal{max}}$, $p_{\textnormal{base}}$} \\
\SetKwFunction{FVAE}{VAE\_Encode}
\SetKwFunction{FFORWARD}{ForwardDiffusion}
$\V{z}_0 \leftarrow$ \FVAE{$\mathcal{I}_{p}$} \\
$\V{z}_{T_b} \leftarrow$ \FFORWARD{$\V{z}_0, T_b$}

\end{algorithm}

%%%%%%%%%%%%

% \setlength{\floatsep}{0.em}
% \setlength{\textfloatsep}{0.2em}
\begin{algorithm}[!h]
\caption{Adaptive joint diffusion}
\label{algo:ajd}
{
    \footnotesize
    \textbf{Input:} $\V{z}_{T_b} \in \mathbb{R}^{H_z \times W_z \times C_z}$ (initial noisy latent), $T_b$ (initial timestep), $H_x$ (height of latent view), $W_x$(height of latent view), $s_{\text{back}}$ (stride in background region), $s_{\text{inst}}$ (stride in instance region), $\beta_{\text{over}}$ (overlap threshold), $\V{p}_{\text{inst}}$ (entire pose map), ${\C{M}}_{\text{inst}}$ (list of instance masks), $\C{Y}_{\text{inst}}$ (list of instance texts),  $y_{\text{global}}$ (global text)

    \textbf{Output:} $ \V{z}_{0}$ (output latent)
    \SetKwProg{Fn}{Function}{:}{}

    \SetKwFunction{FGVAS}{GetViews\_AdaptStride}
    \SetKwFunction{FGIAC}{GetInputs\_AdaptConds}
    % \SetKwFunction{FDiffStep}{DiffusionStep}
    \medskip
    \bluetext{ \tcp{Get view coordinates with adaptive stride}}
    $\C{V}, N_{\text{view}} \leftarrow \FGVAS(H_z, W_z, H_x, W_x, s_{\text{back}},s_{\text{inst}}, \C{M}_{\text{inst}}, \beta_{\text{over}}$) \\
    \medskip
    \bluetext{ \tcp{Diffusion loop}}
    \For{ $t =  T_b,\dots,1$}{
        \bluetext{ \tcp{Set-up count variable and denoised latent}}
        $\V{c} \in \mathbb{R}^{H_z \times W_z} \leftarrow \bm{0}$ \\
        $\V{z}_{t-1} \in \mathbb{R}^{H_z \times W_z \times C_z} \leftarrow \bm{0}$ \\ 
        \For{$i =  0,\dots,N_{\textnormal{view}}-1$}{
            \bluetext{ \tcp{Get inputs of diffusion model with adaptive conditioning}}
            \scriptsize{$\V{x}^{(i)}_t, \V{p}_{\text{view}}, y_{\text{view}} \leftarrow \FGIAC(\C{V}^{(i)}, \V{z}_t, \V{p}_{\text{inst}}, \C{M}_{\text{inst}}, \C{Y}_{\text{inst}}, y_{\text{global}})$} \\
            \footnotesize
            \medskip
            \bluetext{ \tcp{Take a diffusion sampling step for each view}}
            $\V{x}^{(i)}_{t-1} \leftarrow \mathcal{D}(\V{x}^{(i)}_t, \V{p}_{\text{view}}, y_{\text{view}}$)  \\ 

            \medskip
            \bluetext{ \tcp{Fill-up count variable and denoised latent}}
            $h_1, h_2, w_1, w_2 \leftarrow  \C{V}^{(i)}$  \\
            $\V{c}^{(h_1: h_2, w_1: w_2)} \leftarrow \V{c}^{(h_1: h_2, w_1: w_2)} + \bm{I}  $  \\ 
            $\V{z}_{t-1}^{(h_1: h_2, w_1: w_2)} \leftarrow \V{z}_{t-1}^{(h_1: h_2, w_1: w_2)} + \V{x}^{(i)}_{t-1}$  \\ 
        }
        \medskip
        \bluetext{ \tcp{Average overlapped region}}
        $\V{z}_{t-1} \leftarrow \V{z}_{t-1}/\V{c}$ \\
        }
}

\end{algorithm}

\setlength{\floatsep}{0.em}
\setlength{\textfloatsep}{0.2em}
\begin{algorithm}[!h]
\caption{Adaptive conditioning}
\label{algo:ac}
{
    \footnotesize
    \SetKwProg{Fn}{Function}{:}{}

    \SetKwFunction{FGIAC}{GetInputs\_AdaptConds}
    % \bluetext{\scriptsize \tcp{Denoise each view-wise latent}}
    \Fn{\FGIAC{$\V{v}, \V{z}_t, \V{p}_{\text{inst}}, \C{M}_{\text{inst}}, \C{Y}_{\text{inst}}, y_{\text{global}}$}}{
        $h_1, h_2, w_1, w_2 \leftarrow  \V{v}$ \\
        \medskip
        \bluetext{ \tcp{Get input pose map and latent for each view}}
        $\V{p}_{\text{view}} \leftarrow  \V{p}_{\textnormal{inst}}^{(h_1: h_2, w_1: w_2)}$ \\
        $\V{x}_t  \leftarrow \V{z}_t^{(h_1: h_2, w_1: w_2)}$ \\
        \medskip
        \bluetext{ \tcp{Get input text for each view}}
        $y_{\text{view}} \leftarrow  ``\ "$ \\
        \For{$i = 0, \dots N_{\textnormal{inst}}$}{
            $y, \V{m} \leftarrow  \C{Y}_{\text{inst}}^{(i)}, \C{M}_{\text{inst}}^{(i)}$ \\
            \If{ $\sum^{h_2-1}_{k=h_1} \sum^{w_2-1}_{l=w_1} \V{m}^{(k,l)} > 0$}{
                $y_{\text{view}} \leftarrow  y_{\text{view}} + y$ \\
            }       
        }
        \If{ $y_{\textnormal{view}}  = ``\ "$}{
                $y_{\text{view}}  \leftarrow y_{\text{global}}$ \\
            }
        \KwRet $\V{x}_t, \V{p}_{\text{view}}, y_{\text{view}}$
    } 
}

\end{algorithm}

\setlength{\floatsep}{0.em}
\setlength{\textfloatsep}{0.2em}
\begin{algorithm}[!h]
\caption{Adaptive stride}
\label{algo:as}
{
    \footnotesize
    \SetKwProg{Fn}{Function}{:}{}

    \SetKwFunction{FGVAS}{GetViews\_AdaptStride}
    % \bluetext{\scriptsize \tcp{Denoise each view-wise latent}}
    \Fn{\FGVAS{$H_z, W_z, H_x, W_x, s_{\text{back}},s_{\text{inst}}, \C{M}_{\text{inst}}, \beta_{\text{over}}$}}{
        \bluetext{ \tcp{Compute total instance mask and stride ratio}}
        $\V{m}_{\text{total}} \leftarrow \bigcup_i  \C{M}_{\text{inst}}^{(i)}$ \\
        $r_{\text{str}} \leftarrow s_{\text{inst}}//s_{\text{back}}$ \\
        \medskip
        \bluetext{ \tcp{Compute the  default number of views}}
        $N_h \leftarrow (H_z - H_x) // s_{\text{back}} + 1 $ ; $N_w \leftarrow (W_z - W_x) // s_{\text{back}} + 1 $ \\
        $N_{\text{view}} \leftarrow N_h \cdot N_w $ ; $N'_{\text{view}} \leftarrow N_{\text{view}} $\\
        $\C{V} \leftarrow [\ ]$ \\
        \medskip
        \For{$i = 0, \dots, N_{\textnormal{view}}-1$} {
             \bluetext{ \tcp{Get a default view coordinate}}
             $h_1 \leftarrow (i // N_w) \cdot s_{\text{back}}$ ; $h_2 \leftarrow h_1 + H_x$  \\ 
             $w_1 \leftarrow (i \% N_w) \cdot s_{\text{back}}$ ; $w_2 \leftarrow w_1 + W_x$  \\ 
             Append $(h_1, h_2, w_1, w_2)$ to $\C{V}$\\
             \medskip
             \bluetext{ \tcp{Compute overlap ratio}}
             $r_{\text{over}} = \frac{\sum^{h_2-1}_{k=h_1} \sum^{w_2-1}_{l=w_1} \V{m}_{\text{total}}^{(k,l)}}{H_z\cdot W_z}$\\
             \medskip
             \bluetext{ \tcp{Get view coordinates with adaptive stride}}
             \If{$r_{\textnormal{over}} > \beta_{\textnormal{over}}$}{
                \For{$ j=1, \dots, r_{\textnormal{str}}^2$}{
                    $h_1 \leftarrow h_1 + (j//r_{\text{str}})\cdot s_{\text{inst}} $ ; $h_2 \leftarrow h_1 + H_x$  \\ 
                    $w_1 \leftarrow w_1 + (j\%r_{\text{str}})\cdot s_{\text{inst}} $ ;  $w_2 \leftarrow w_1 + W_x$  \\ 
                    Append $(h_1, h_2, w_1, w_2)$ to $\C{V}$ \\
                    $N'_{\text{view}} \leftarrow N'_{\text{view}}+1$ \\
                    }
                }
        }
        \KwRet $\C{V}, N'_{\text{view}}$
    } 
}

\end{algorithm}

\subsubsection{Adaptive joint diffusion}

The adaptive joint process aims to generate a clean, high-resolution latent variable, $\V{z}_{0}$, from a noisy latent variable, $\V{z}_{T_b} \in \mathbb{R}^{H_z \times W_z \times C_z}$, injected with high frequencies at timestep $T_b$. As detailed in Algorithm~\ref{algo:ajd}, the process begins by obtaining a list of view coordinates, $\C{V}$, using an \textit{adaptive stride} for efficient joint diffusion. We then iteratively denoise the latent variable. At each timestep, each view of latent $\V{x}^{(i)}_{t}$ is denoised using \textit{adaptive conditioning} on its corresponding input latent $\V{x}^{(i)}_{t}$, pose map $\V{p}_{\text{view}}$, and text $y_{\text{view}}$.
This leverages information specific to each view to resolve the issue of duplicated objects. Following this conditioning, a diffusion sampling step (as described in~\cite{ho2020denoising, song2020denoising, liu2022pseudo}) is applied to each view. The denoised views are combined by averaging overlapping regions to form the next timestep's latent variable, $\V{z}_{t-1}$. This process continues iteratively until the final clean latent variable, $\V{z}_{0}$, is obtained.
More details in the adaptive conditioning and adaptive stride are provided below.

\paragraph{Adaptive conditioning}
The process incorporates adaptive view-wise conditioning to ensure each view incorporates relevant text and pose information as represented in Algorithm~\ref{algo:ac}.
For each view, we extract the corresponding input pose $\V{p}_{\text{view}}$, and latent code $\V{x}_t$, by cropping from the entire pose map $\V{p}_{\text{inst}}$, and latent variable $\V{z}_t$. We then utilize the instance mask $\V{m}$, to determine which human instances are present within the view. If a view contains an instance, its corresponding text description $y$, is included into the view's text input $y_{\text{view}}$ . Conversely, views without human instances solely rely on the global text description $y_{\text{global}}$.
Furthermore, if the detailed text description mentions specific body parts (head, face, upper body, etc.), these details can be applied to corresponding regions using fine-grained segmentation maps. This approach promotes efficient and robust joint diffusion while granting control over critical human characteristics like pose and appearance.

\paragraph{Adaptive stride}
To capture finer details, particularly in the areas with human instances, the joint process employs an adaptive stride as represented in Algorithm~\ref{algo:as}. We first calculate a total instance mask, $\V{m}_{\text{total}}$, where all areas containing instances are marked as 1 and the background is marked as 0. We then define a stride ratio $r_{\text{str}}$, between the instance stride $s_{\text{inst}}$, and the background stride $s_{\text{back}}$.
Next, we determine the number of views based on the entire latent size $H_z \times W_z$ the individual view size $H_x \times W_x$, and the background stride $s_{\text{back}}$.
We then populate the total view list, $\C{V}$, with default view coordinates for each view. To ensure capturing fine details in human instance regions, we calculate an overlap ratio $r_{\text{over}}$, which represents the proportion of a view that contains human instances. If this overlap ratio exceeds a threshold $\beta_{\text{over}}$, we add additional views corresponding to the finer instance stride $s_{\text{inst}}$. This ensures denser sampling in these areas for a more detailed reconstruction of the scene.

%%%%%%%%%%%%%%%%%%%%%%%%%%%%%%%%%%%%%%%%%%%%%%%%%%%
\section{Implementation Details of Our Proposed Method}

\subsection{Detailed Base Image Generation }
Our approach leverages several techniques for efficient human instance generation and integration within the scene. First, we utilize SDXL-ControlNet-Openpose~\cite{podell2023sdxl, controlnet23, sdxl_controlnet_openpose}  to directly generate human instances based on text descriptions and pose information.
For accurate human segmentation, we employ Lang-Segment-Anything~\cite{lang_sam}, a language-conditioned segmentation model. This model efficiently extracts human regions from base images using prompts like ``person'' or ``human''. 
For specific human parts segmentation, we first separate the head region into ``head'' and ``hair'' using the same model, and then combine them to form the head segmentation. We then perform segmentation on the body parts, which consist of the entire human body except for the head segmentation.
Subsequently, we optionally re-estimate human poses within the generated images using two models trained on whole-body pose datasets:  ViTPose~\cite{xu2022vitpose} and YOLOv8 detector~\cite{yolov8}.
Finally, for seamless integration of the foreground elements with the background, we first resize and create a base collage. SDXL-inpainting~\cite{podell2023sdxl, sdxl_inpainting} is then employed to inpaint the generated foreground elements onto the background.  To handle backgrounds of arbitrary sizes, we implement joint diffusion~\cite{bar2023multidiffusion} with SDXL-inpainting.
%We employ a stride value of 32 and a window size of 128 during inpainting.
% 

\subsection{Instance-Aware Hieracical Enlargement}

\subsubsection{High frequency-injected forward diffusion}
We implement the Canny edge detection algorithm with thresholds of 100 and 200. To smooth the edge map, a Gaussian kernel with a standard deviation $\sigma$ of 50 is used. The probability map $\mathcal{C}$ is constructed by normalizing and conditioning the blurred edge map. We define a high probability threshold  $p_{\textnormal{max}}$ of 0.1 and a low probability threshold $p_{\textnormal{base}}$ of 0.005. Lanczos interpolation is employed for image upscaling. $d_r$ and $\alpha_{interp}$ is set to 4 and 2 respectively, for pixel perturbation based on probability map. Finally, the forward diffusion timestep $T_b$ is set to 700, which is 0.7 times the total training steps of 1000 used in the SDXL framework.

\subsubsection{Adaptive joint process}
For the adaptive joint process, which receive the generated pose map and high frequency-injected noisy latent as input, SDXL-ControlNet-Openpose~\cite{podell2023sdxl, controlnet23, sdxl_controlnet_openpose} is employed. When using an adaptive stride, $\beta_{over}$ is set to 0.2, the background stride $s_{back}$ is set to 64 and $s_{inst}$ is set to 32. When adaptive stride was not employed, both $s_{back}$ and $s_{inst}$ were set to 32.

\begin{figure}[!h]
    \centering
    \includegraphics[width=0.99\textwidth]{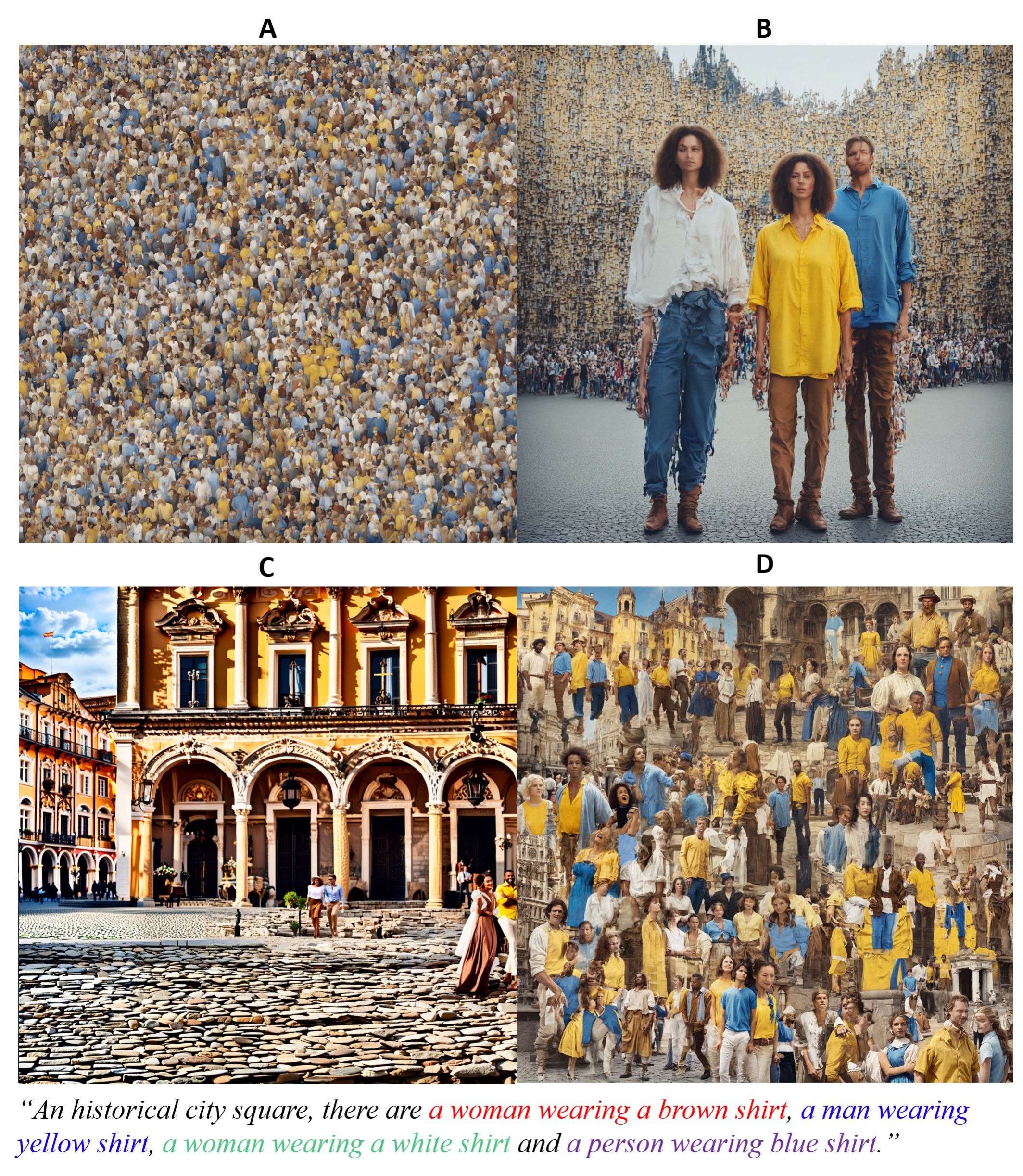}
  % \vspace{-1em}
    \caption{User study comparing high-resolution images (4096$\times$4096) generated from detailed text descriptions. Participants ranked the four anonymized images (A, B, C, D) based on three criteria: text-to-image correspondence, overall image naturalness, and human naturalness. In this example, Model A corresponds to ControlNet~\cite{podell2023sdxl, controlnet23}, Model B to ScaleCrafter~\cite{he2023scalecrafter}, Model C to BeyondScene (ours), and Model D to MultiDiffusion~\cite{bar2023multidiffusion}. The order of models was shuffled during the study.}
    \label{fig_user}
  % \vspace{-1em}
\end{figure}

%%%%%%%%%%%%%%%%%%%%%%%%%%%%%%%%%%%%%%%%%%%%%%%%%%%
\section{Details on User Study}
% (동운) Details + Exmaple 광장
We employed a crowd sourcing for a user study that evaluate the \textit{text-image correspondence} and \textit{naturalness}. We presented participants with high-resolution images generated by SDXL~\cite{podell2023sdxl}, MultiDiffusion~\cite{bar2023multidiffusion}, ScaleCrafter~\cite{he2023scalecrafter}, and our BeyondScene. The guidelines for ranking the methods by participants are that rank the generated images in order of (1) their \textit{text-image correspondence} focusing on how accurately the images captured all the elements from the text prompts without duplication or missing instances, (2) \textit{global naturalness}, particularly regarding physically impossible elements, disconnected objects, and overall background coherence, and (3) \textit{human naturalness}, specifically focusing on anatomical anomalies like unusual facial features (eyes, nose and mouth, etc.), hands, feet, legs, and overall body structure. We shuffled the order of images and randomly selected captions from CrowdCaption dataset. We presented the generated images from 4 results vertically. In our user study, 101 participants completed the survey, contributing a total of 12,120 votes. In the Fig.~\ref{fig_user},  we present an illustrative example from the user study that evaluated images generated by SDXL~\cite{podell2023sdxl}, MultiDiffusion~\cite{bar2023multidiffusion}, ScaleCrafter~\cite{he2023scalecrafter}, and our BeyondScene.

\setlength{\floatsep}{12pt}
\setlength{\textfloatsep}{20pt}

%%%%%%%%%%%%%%%%%%%%%%%%%%%%%%%%%%%%%%%%%%%%%%%%%%%

\begin{figure}[!t]
    \centering
    \includegraphics[width=0.99\textwidth]{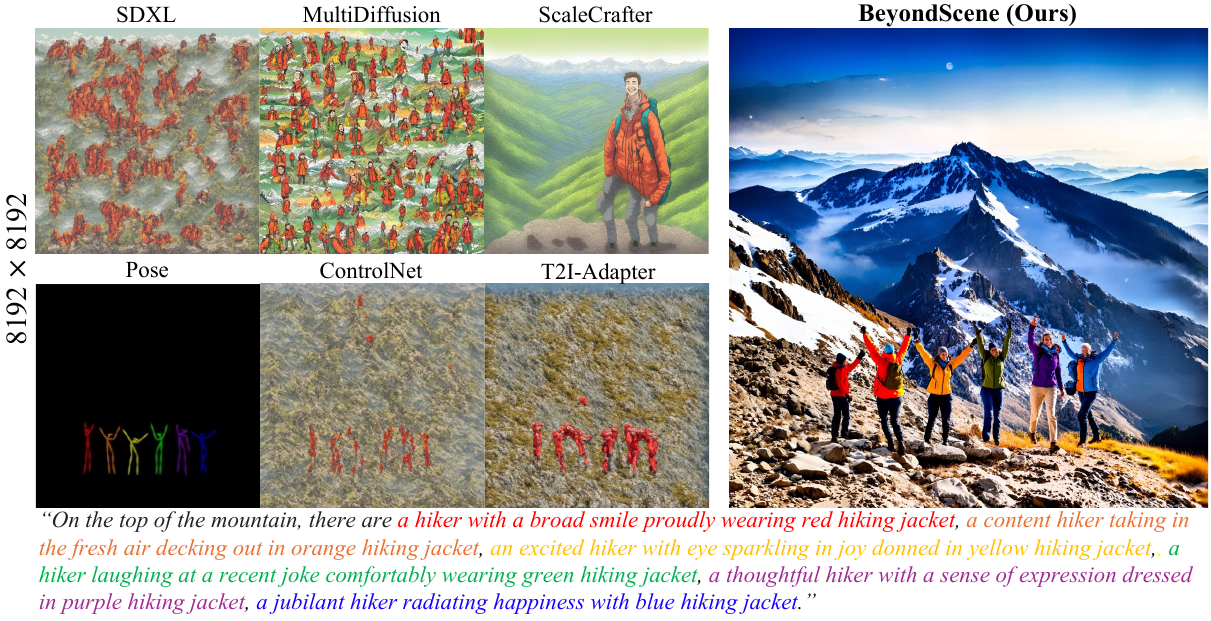}
  % \vspace{-1em}
    \caption{Qualitative comparison between baselines and our BeyondScene in $8192\times 8192$ resolution. The color in each description represents
the description for each instance that has the same color in the pose map. BeyondScene succeed in generating high-fidelity results with great text-image correspondence, while other baselines fail.}  
    \label{fig_8k}
  % \vspace{-1em}
\end{figure}

\section{Additional Results}
\subsection{Comparison at 8192$\times$8192 Resolution}
BeyondScene, our high-resolution human-centric scene generation framework, achieves exceptional resolution scalability, generating images beyond 8K resolution. We conducted a qualitative evaluation (Fig.~\ref{fig_8k}) to assess BeyondScene's capability for ultra-high resolution image generation. As seen in the upper image of Fig.~\ref{fig_8k}, where baselines struggle to accurately reflect prompts or depict proper human anatomy, BeyondScene produces clear, high-fidelity images that faithfully adhere to all prompts, including details like the order of clothing colors. Specifically, methods like SDXL~\cite{podell2023sdxl}, MultiDiffusion~\cite{bar2023multidiffusion}, and ScaleCrafter~\cite{he2023scalecrafter}, which solely rely on text input, generate overly-duplicated instances or exhibit anatomical inconsistencies. In contrast, BeyondScene demonstrates high-fidelity results with accurate grounding in human anatomy. Notably, while baselines like ControlNet~\cite{podell2023sdxl, controlnet23} and T2IAdapter~\cite{podell2023sdxl,t2i23} that leverage visual priors fail to generate a person based on the provided pose map, BeyondScene successfully generates images that precisely follow the pose guide, resulting in superior quality.

\subsection{Efficiency Analysis}
To assess BeyondScene's computational efficiency, we compared its GPU peak memory usage and floating-point operations (FLOPs) to existing methods~\cite{bar2023multidiffusion, lee2023syncdiffusion, he2023scalecrafter} in Tab.~\ref{tab:efficiency}. BeyondScene achieves favorable efficiency: it utilizes similar or fewer FLOPs than the joint diffusion approach while maintaining similar or lower memory usage. Compared to the dilation-based method~\cite{he2023scalecrafter}, BeyondScene demonstrably requires less memory. Also, it's important to note that for both Regional-MultiDiffusion~\cite{bar2023multidiffusion} and BeyondScene, the computational cost scales with the number of instances requiring grounding. Specifically, MultiDiffusion~\cite{bar2023multidiffusion} incurs an additional 36.8 PFLOPs per instance, while BeyondScene requires only 2.59 PFLOPs. Furthermore, unlike MultiDiffusion, BeyondScene maintains its peak memory usage regardless of the number of instances.

In Table~\ref{tab:ada_stride}, we showcase the efficiency gains of adaptive stride. Compared to a fixed stride of 32, our approach significantly reduces computational cost (measured in PFLOPs). Furthermore, qualitative comparisons in Figure~\ref{fig_ablation_adaptive_stride} reveal a trade-off between computational cost and image quality with fixed strides. While a stride of 64 offers lower cost, it introduces unnatural anatomy in the human figure, even though the background appears acceptable.  BeyondScene with adaptive strides, however, achieves both the detail and coherence of a stride-32 model, without the anatomical artifacts, all at a lower computational cost.

\begin{table}[t]
\caption{Comparison of efficiency with GPU memory usage and floating point operations (FLOPs) between baselines and ours; MultiDiffusion (Multi.)~\cite{bar2023multidiffusion}. Regional-MultiDiffusion (R-Multi.)~\cite{bar2023multidiffusion}, SyncDiffusion (Sync.)~\cite{lee2023syncdiffusion}, ScaleCrafter (Scale.)~\cite{he2023scalecrafter}, and BeyondScene. All the images generation resolution for the evaluation was set to $4096\times 4096$, and all the stride of the joint diffusion-based prior works was set to 32, while BeyondScene was implemented with adaptive stride which has less views than the others. The number of instance for the image generation was set to 3.}
\label{tab:efficiency}
\resizebox{\textwidth}{!}{%
\begin{tabular}{cccccc}
\specialrule{.1em}{.05em}{.05em}
                  & Multi.~\cite{bar2023multidiffusion} & R-Multi.~\cite{bar2023multidiffusion} & Sync.~\cite{lee2023syncdiffusion} & Scale.~\cite{he2023scalecrafter} & BeyondScene (Ours) \\ \hline
Memory usage (GB) & 19.695         & 32.178           & 64.049        & 28.158       & 26.992             \\
PFLOPs            & 73.5           & 553              & 221           & 10.6         & 98.7               \\
\specialrule{.1em}{.05em}{.05em}
\end{tabular}%
}

\end{table}

\begin{table}[!t]
\centering
\caption{Computational cost (PFLOPs) of our BeyondScene as the stride size varies. Adaptive stride requires fewer PFLOPs compared to using a full 32 stride. The computation costs were measured by generating a high-resolution human-centric scene (4096$\times$4096) containing 3 instances.}
\begin{tabular}{cccc}
\specialrule{.1em}{.05em}{.05em}
Stride size           & 32             & 64               & Adaptive stride             \\ \hline
PFLOPs            &\;\;\; 119.3 \;\;\;          & \;\;\; 17.7 \;\;\;              & \;\;\;98.7\;\;\;                          \\
\specialrule{.1em}{.05em}{.05em}
\end{tabular}
\label{tab:ada_stride}
\end{table}

\begin{figure}[!t]
    \centering
    \includegraphics[width=0.99\textwidth]{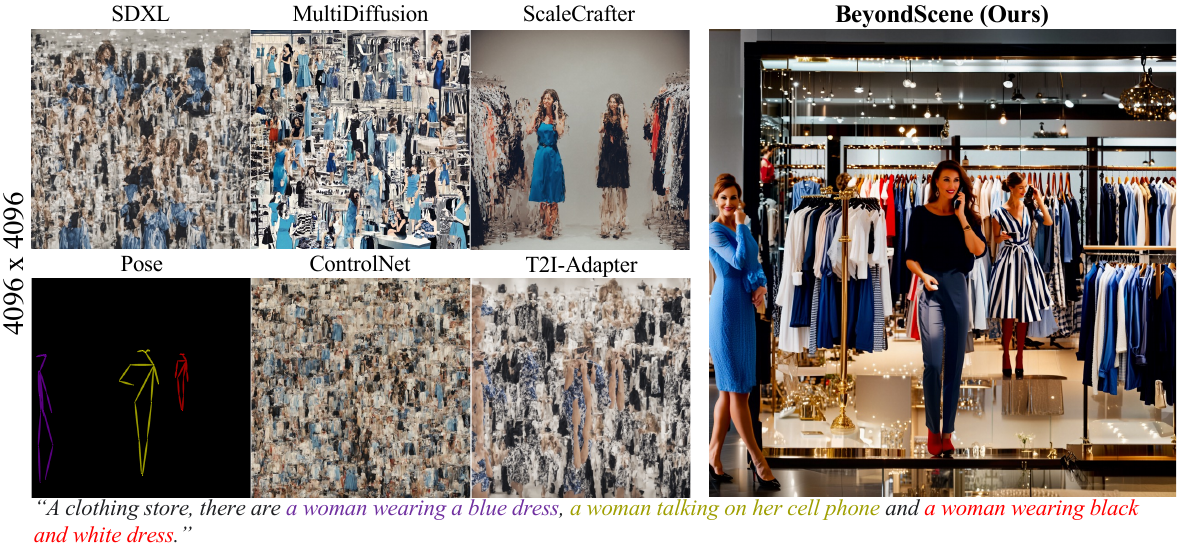}
  % \vspace{-1em}
    \caption{Additional examples of large scene synthesis (4096$\times$4096) on the poses and text obtained from CrowdCaption~\cite{wang2022happens} images. 
    All baselines including SDXL~\cite{podell2023sdxl}, MultiDiffusion~\cite{bar2023multidiffusion},  ScaleCrafter~\cite{he2023scalecrafter}, ControlNet~\cite{controlnet23}, and T2IAdapter~\cite{t2i23}) produce duplicated objects and artifacts in human anatomy, while our method succeeded in generation of high-resolution image with high text-image correspondence. Each color in the description corresponds to instances sharing the same color in the pose map.}  
    \label{fig_quali_crowd_square}
  % \vspace{-1em}
\end{figure}

\begin{figure}[!h]
    \centering
    \includegraphics[width=0.99\textwidth]{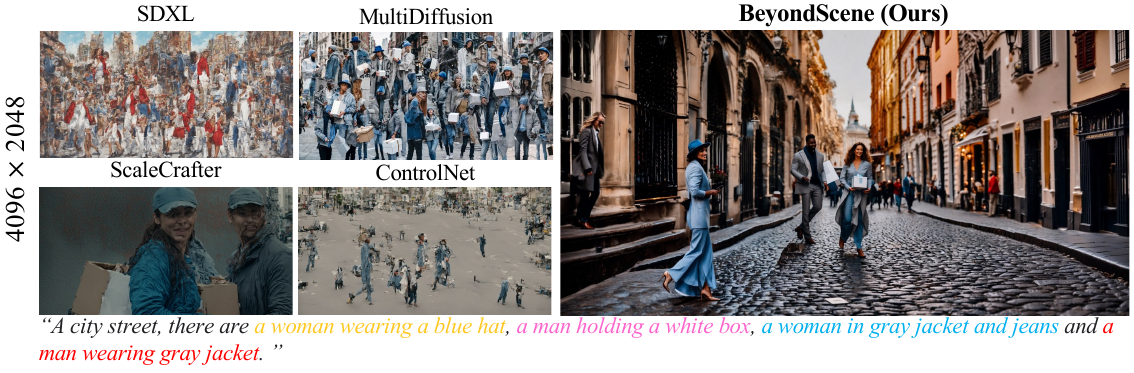}
  % \vspace{-1em}
    \caption{Additional examples of large scene synthesis (4096$\times$2048) on the poses and text obtained from CrowdCaption~\cite{wang2022happens} images. Compared to existing approaches like SDXL~\cite{podell2023sdxl}, MultiDiffusion~\cite{bar2023multidiffusion}, ScaleCrafter~\cite{he2023scalecrafter}, and ControlNet~\cite{podell2023sdxl, controlnet23}, our method achieves minimal artifacts, strong text-image correspondence, and high global and human naturalness.}  
    \label{fig_quali_crowd_wide}
  % \vspace{-1em}
\end{figure}
\subsection{Additional Qualitative Comparison}
%(하연,회기,동운) Custom Example, CrowdCaption, 
For qualitative comparisons, we present the additional examples of generated high-resolution human-centric scenes from CrowdCaption dataset~\cite{wang2022happens}. As shown in the Fig.~\ref{fig_quali_crowd_square} and ~\ref{fig_quali_crowd_wide}, the baselines introduce the duplication issue and unnatural human anatomy artifacts, thereby suffering from low text-image correspondence and naturalness. In contrast, our method achieves a high level of detail in human characteristics while matching the number of humans present. It also retains the scene's overall naturalness by reflecting real-world physics and minimizing artifacts related to human anatomy, closely adhering to the given texts.

% \begin{figure}[!t]
  % \vspace{-1em}
   %     \centering
   %  \includegraphics[width=0.99\textwidth]{figures/control.pdf}
   % \vspace{-1em}
   %  \caption{Qualitative results for effectiveness of our method for progressively added detailed descriptions of each instance. Our method succeeds in generating high-resolution human scenes with strong text-image correspondence for added detailed descriptions.}        
   %  \label{fig:ablation_detailed_text}
% \end{figure}

% \subsubsection{Adding detailed text} 
% As represented in Fig.~\ref{fig:ablation_detailed_text}, with increasing detail in the text description, the generated image progressively evolves in a controlled manner, visually reflecting each added nuiance.

\subsection{Generation Beyond Token Limits} %(Figure reference tentative)}
Our BeyondScene method surpasses the token limitations of existing diffusion models, enabling the generation of richer and more detailed instances. For example, in Fig.1 (main paper), BeyondScene handles text prompts exceeding the 77-token limit of SDXL, with 99 and 128 tokens, respectively, resulting in significantly more detailed generations for various instances in the figure. Similarly, in Fig.5 (upper, main paper), BeyondScene effectively utilizes a 205-token input to capture the unique characteristics of each ballerina. Because of the shortage of space in main paper, we shorten the input prompt for readability and the full prompt to generate the image is in Tab.~\ref{tab:full_prompt}. The image of Fig.~\ref{fig_8k} exceeds the limit of CLIP text encoder with 102-token. Also, Fig.~\ref{fig_quali_custom_3584} (beach, campfire) and Fig.~\ref{fig_quali_custom_7186} (Alps) showcase successful generation with inputs exceeding 77 tokens (83, 109, and 104-tokens, respectively).

\begin{table}[h!]
\caption{Full text captions used to generate images in the main paper, exceeding the 77-token limit of the pre-trained diffusion model.}
\label{tab:full_prompt}
\begin{tabular}{ p{\dimexpr 0.1\linewidth}
                 p{\dimexpr 0.9\linewidth}}
\hline
\multicolumn{2}{c}{{Full text captions used to generate images in the main paper}}                \\ \hline
Fig.1 (Upper) &  In the background garden of 3D game, there are / girl in red dress, Zelda character, wearing red dress with traditional leather accessories and colorful patterns / girl in blue dress, Zelda  character,  wearing blue dress with traditional leather accessories and colorful patterns /  girl in green dress, Zelda character,  wearing green dress with traditional leather accessories and colorful patterns / girl in yellow dress, Zelda character,  wearing yellow dress with traditional leather accessories and colorful patterns. \newline \\
Fig.1 (Lower) & In the the background of sunny road, there are / a clay animation style character with spiky hair and pair of goggles on its head and sports a mischievous grin, / a clay animation style character with short, chubby cheeks and a round body with oversized glasses, / a clay animation style old man character with tall, slender frame and two eyes of equal size with scientist white lab coat, / a clay animation style character with brown hair, chubby cheek with blue jeans on it, / a clay animation style character with black hair wearing yellow t-shirts, / a clay animation style character with cute red dotted dress and oversize glasses. \newline \\
Fig.5 (Upper) & In the background of the empty stage in opera house, there are / a dancer in a pink ballet suit is doing ballet with sparkling cubics and silver accessories / a dancer in a light blue ballet suit is doing balle. with sparkling cubics and silver accessories / a dancer in a pink ballet suit is doing ballet with sparkling cubics and silver accessories / a dancer in a light blue ballet suit is doing balle. with sparkling cubics and silver accessories / a dancer in a yellow ballet suit is doing ballet. wearing big silver tiara on her head with sparkling cubics, silver accessories and necklace / a dancer in a light blue ballet suit is doing ballet. with sparkling cubics and silver accessories / a dancer in a pink ballet suit is doing ballet with sparkling cubics and silver accessories / a dancer in a light blue ballet suit is doing balle. with sparkling cubics and silver accessories / a dancer in a pink ballet suit is doing ballet with sparkling cubics and silver accessories.
\\
\hline
\end{tabular}

\end{table}

\subsection{Reasoning of MLLM-Based Metrics on Human-Centric Scene}
\begin{figure}[]
    \centering
    \includegraphics[width=0.85\textwidth]{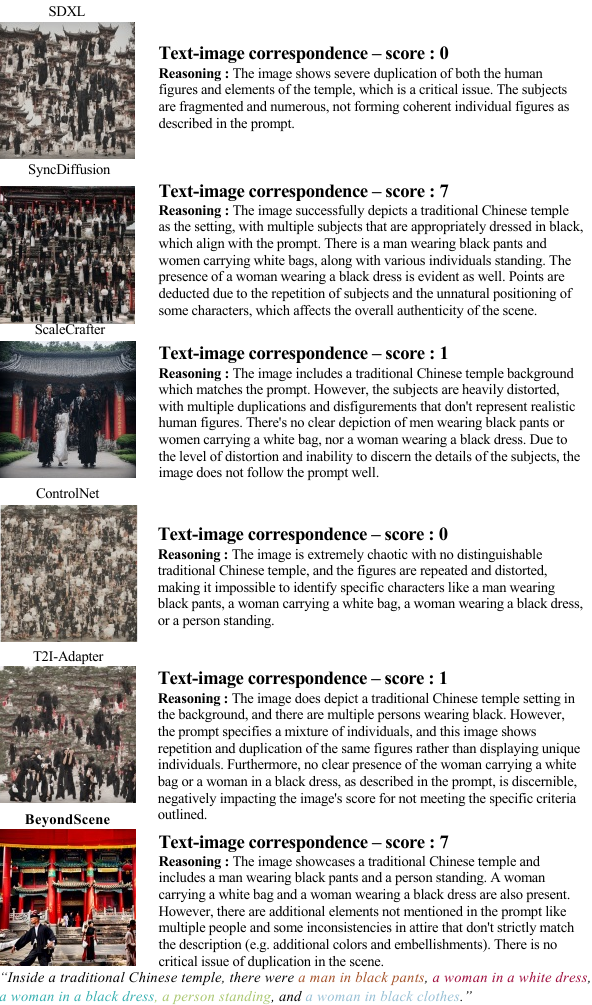}
  % \vspace{-1em}
    \caption{Examples of {text-image correspondence} score from VIEScore~\cite{ku2023viescore}, powered by the GPT-4~\cite{achiam2023gpt} multimodal language model (MLLM) (4096$\times$4096). The MLLM assigns a {text-image correspondence} score between 0 and 10, along with a detailed reasoning for the score.}  
    \label{fig_text_2_image_corr}
  % \vspace{-1em}
\end{figure}

\begin{figure}[]
    \centering
    \includegraphics[width=0.85\textwidth]{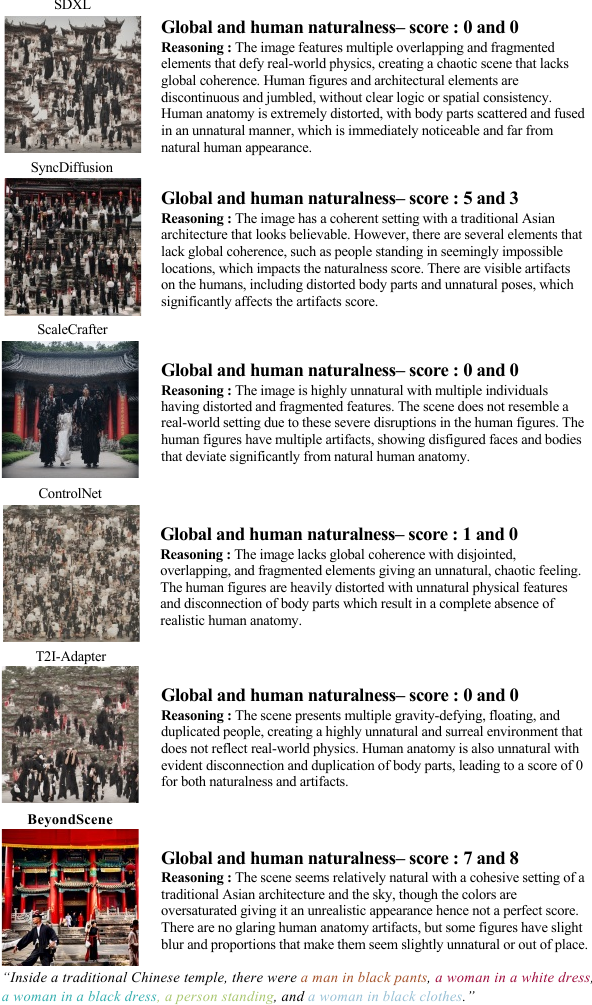}
  % \vspace{-1em}
  \caption{Examples of {global naturalness} and {human naturalness} scores from VIEScore~\cite{ku2023viescore}, powered by the GPT-4~\cite{achiam2023gpt} multimodal language model (MLLM) (4096$\times$4096). The MLLM assigns a {global naturalness} and {human naturalness} between 0 and 10, along with a detailed reasoning for the score.}  
    \label{fig_naturalness}
  % \vspace{-1em}
\end{figure}

%(동운) 예제 1개 baseline과 비교
For evaluating generated high-resolution human-centric scenes, we adopt the MLLM-based metric, VIEScore~\cite{ku2023viescore} which leverages a multimodal large language model. Focusing on (1) \textit{text-image correspondence} and (2) \textit{naturalness}, the MLLM-based metric assigns the score from 0 to 10 and provides detailed reasoning. In the Fig.~\ref{fig_text_2_image_corr} and the Fig.~\ref{fig_naturalness}, we visualize the scores and detailed reasoning descriptions from VIEScore~\cite{ku2023viescore} for high-resolution images generated by SDXL~\cite{podell2023sdxl}, SyncDiffusion\cite{lee2023syncdiffusion}, ScaleCrafter~\cite{he2023scalecrafter}, SDXL-ControlNet~\cite{controlnet23}, T2IAdapter~\cite{t2i23}, and our BeyondScene.

\begin{figure}[!t]
    \centering
    \includegraphics[width=0.85\textwidth]{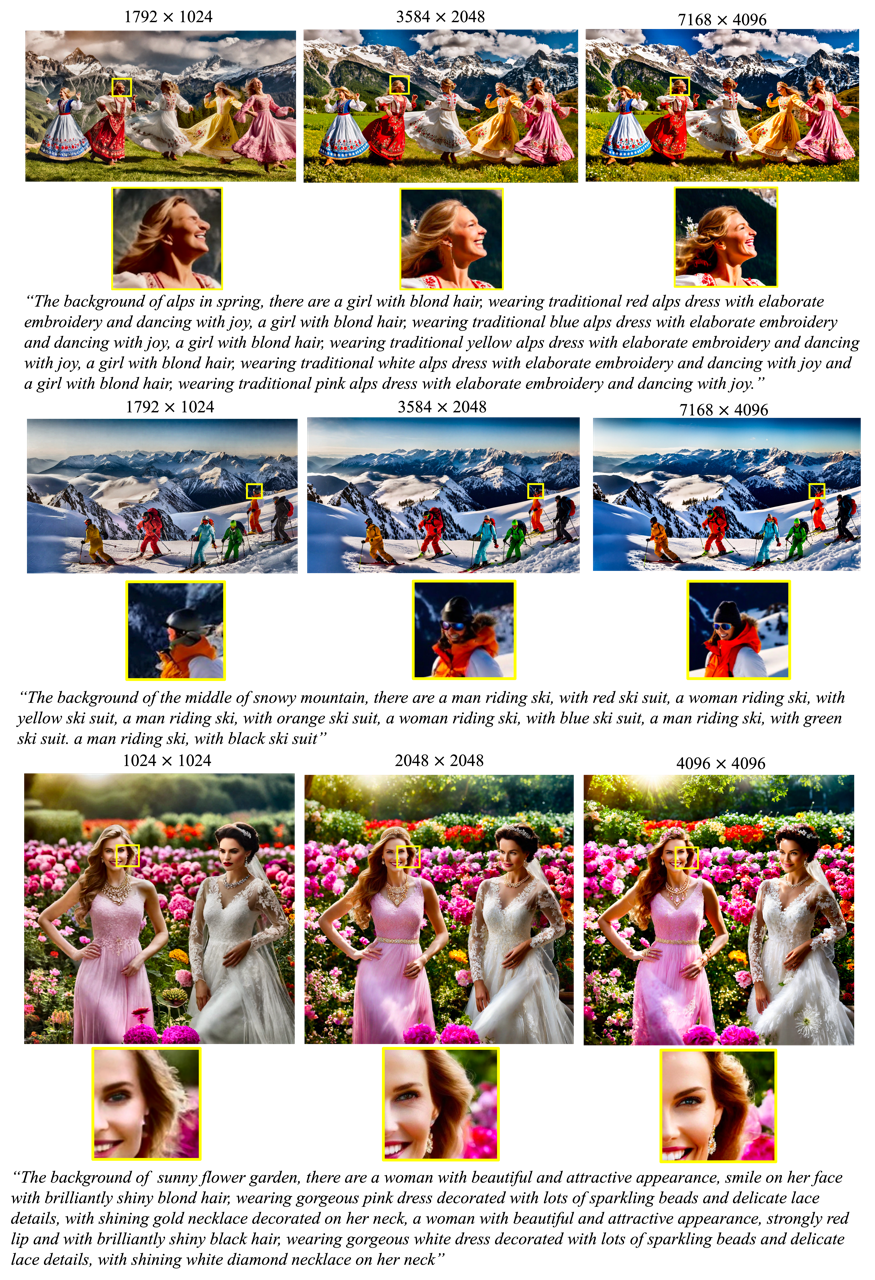}
  % \vspace{-1em}
  \caption{Qualitative results at each stage of our BeyondScene. The leftmost column is the base image generated with collage process. The middle and rightmost columns are images generated with hierarchical enlargement. The hierarchical enlargement progressively refines details like individual hair strands, textures of hats, sunglasses, and intricate designs of earrings, etc.}  
    \label{fig_step}
  % \vspace{-1em}
\end{figure}

\subsection{Results at Each Stage}

As represented in the Fig.~\ref{fig_step}, our BeyondScene gradually improves image resolution and quality from low-resolution to high-resolution through hierarchical enlargement. Due to the collage process, the base image contains slight stylistic and lighting variations between individual instances. These variations are unified through hierarchical enlargement. At each stage of the enlargement process, the details within each instance, like individual hair strands, textures of hats, and designs of earrings, become increasingly distinct.

% \newpage
% \bibliographystyle{splncs04_supp}
% \bibliography{main}

\end{document}